\documentclass[conference]{IEEEtran}
\IEEEoverridecommandlockouts
\usepackage{amsmath,amssymb,amsfonts}
\usepackage{graphicx}
\usepackage{multirow}
\usepackage{textcomp}
\usepackage{bbding}
\usepackage{algorithm}
\usepackage{algorithmicx}
\usepackage{algpseudocode}
\renewcommand\arraystretch{1.25}
\usepackage[normalem]{ulem}
\useunder{\uline}{\ul}{}
\floatname{algorithm}{Algorithm}

\usepackage[numbers,sort]{natbib}
\usepackage{hyperref}

\hypersetup{backref}
\usepackage{balance}
\usepackage{xparse}
\usepackage{multirow}
\usepackage{booktabs}
\usepackage{mathrsfs}
\usepackage{array}
\usepackage{textcomp}
\usepackage{amsthm}
\usepackage{xspace}
\usepackage{float}
\usepackage{xcolor}
\usepackage{bm}
\usepackage{array} 

\begin{document}

\newtheorem{definition}{\textbf{Definition}}

\newtheorem{theorem}{Theorem}
\newtheorem{corollary}{Corollary}[theorem]
\newtheorem{assumption}{Assumption}
\newtheorem{example}{Example}
\newtheorem{lemma}[theorem]{Lemma}

\definecolor{newtext}{rgb}{0, 0, 0}
\newcommand\newtext[1]{\textcolor{newtext}{#1}}

\title{\newtext{Unsupervised Graph Outlier Detection: Problem Revisit, New Insight, and Superior Method}}



\author{
{
Yihong Huang$^\dagger$, Liping Wang$^\dagger$, Fan Zhang$^\ddagger$, Xuemin Lin$^{\S}$}
\vspace{2mm}
\\
\fontsize{10}{10}
\selectfont\itshape
$^\dagger$East China Normal University,
$^\ddagger$Guangzhou University,
$^\S$Shanghai Jiao tong University\\
\fontsize{9}{9} \selectfont\ttfamily\upshape
hyh957947142@gmail.com,
lipingwang@sei.ecnu.edu.cn\\ zhangf@gzhu.edu.cn, 
lxue@cse.unsw.edu.au\\
}
\maketitle

\vspace{-2cm}

\begin{abstract}
A large number of studies on Graph Outlier Detection (GOD) have emerged in recent years due to its wide applications, in which Unsupervised Node Outlier Detection (UNOD) on attributed networks is an important area. 
UNOD focuses on detecting two kinds of typical outliers in graphs: the structural outlier and the contextual outlier. 
Most existing works conduct  experiments based on  datasets with injected outliers. 
However, we find that the most widely-used outlier injection approach has a serious data leakage issue. By only utilizing such data leakage, a simple approach can achieve  state-of-the-art performance in detecting outliers. 
In addition, we observe that existing algorithms have a performance drop \newtext{with the mitigated data leakage issue}. The other major issue is on balanced detection performance between the two types of outliers, which has not been considered by existing studies. 

In this paper, we analyze the cause of the data leakage issue in depth since the injection approach is a building block to advance UNOD. 
Moreover, we devise a novel variance-based model to detect structural outliers, \newtext{which outperforms existing algorithms significantly and is more robust at kinds of injection settings.} On top of this, we propose a new framework, Variance-based Graph Outlier Detection (VGOD), which combines our variance-based model and attribute reconstruction model to detect outliers in a balanced way.  
Finally, we conduct extensive experiments to demonstrate the effectiveness and  efficiency of VGOD. The results on 5 real-world datasets validate that VGOD achieves not only the best performance in detecting outliers but also a balanced detection performance between structural and contextual outliers.
\end{abstract}


\section{Introduction}
Graph Outlier Detection \cite{GOD} (GOD, a.k.a. graph anomaly detection) is a fundamental graph mining task. It has various applications in high-impact domains and complex systems, such as  financial fraudster identification \cite{fraud}. The detection objects of GOD can be classified into different levels like node, edge, community, and graph \cite{god-survey}. For example, detecting abnormal users in a social network is the node-level GOD task while detecting abnormal molecules can be regarded as a graph-level GOD task. 

\par
Due to the high cost or unavailability of manually labeling the ground truth outliers, a large number of existing GOD approaches are carried out in an unsupervised manner \cite{GCNAE,Multi-scale-god}, which aims to detect the instances that significantly deviate from the majority of instances in the graph \cite{deep-ad}. Attributed networks (a.k.a. attributed graphs) are a powerful data representation for many real-world complex systems (e.g. a social network with user profiles), in which entities can be represented as nodes with their attribute information; the interaction or relationship between entities can be represented as edges \cite{attribute-network}. In recent years, the study of Unsupervised Node Outlier Detection (UNOD) on attributed networks has been blooming due to its wide applications \cite{gad-survey-old,god-survey,pygod}.  Different from traditional global outlier detection and time series outlier detection, it defines two new typical types of outliers on attributed networks, namely, structural outlier and contextual outlier \cite{GCNAE}. 

\begin{figure}[!htbp]
		\setlength{\belowcaptionskip}{-0.5cm}   
		\centering
		\includegraphics[width=195.0 pt]{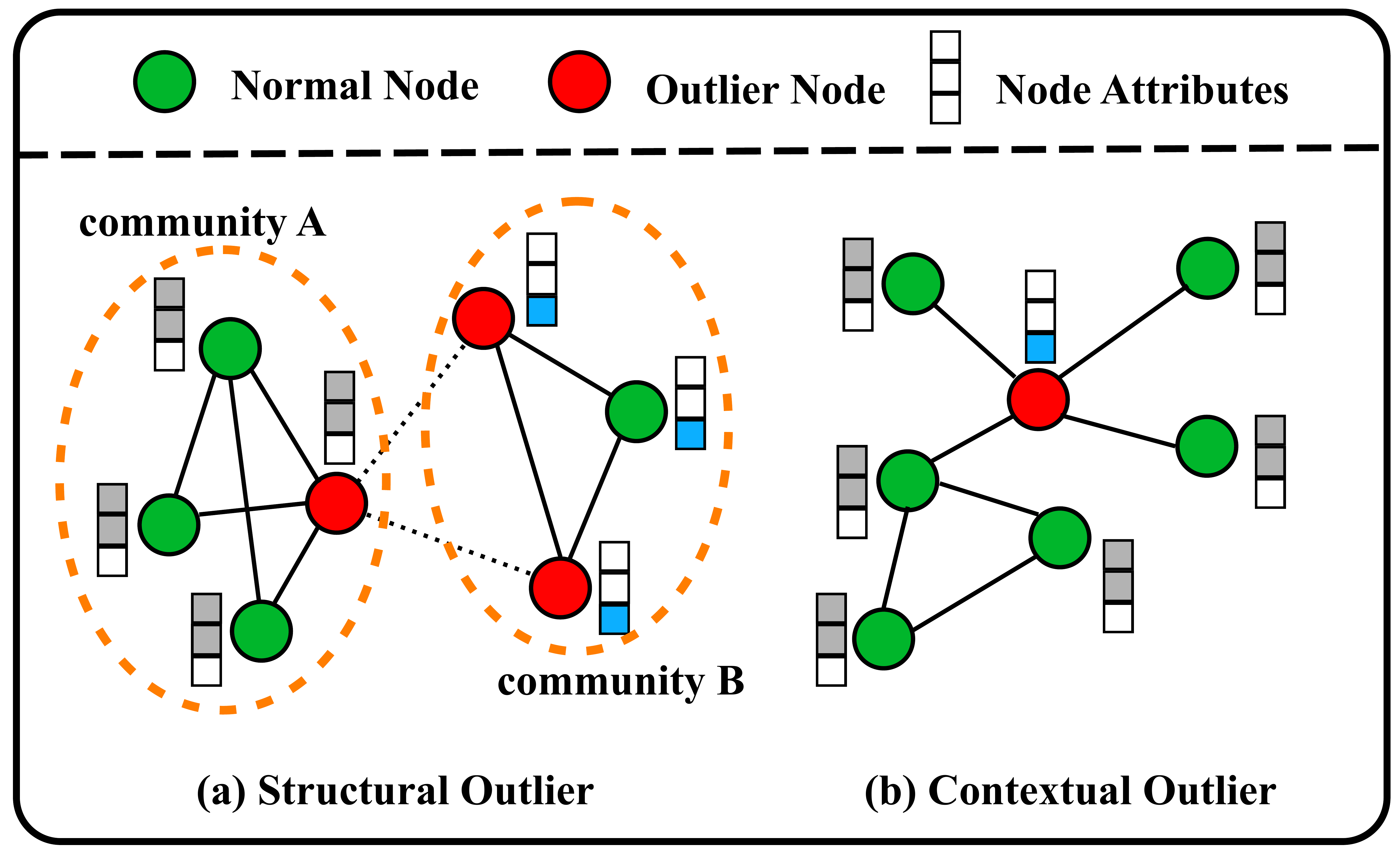}
		\caption{An example of structural and contextual outliers in UNOD.}%
		\label{two-type}
\end{figure}

In Fig \ref{two-type}, there is a toy example for these two kinds of basic outliers. Particularly, structural outliers are those nodes structurally connected to different communities, i.e., their structural neighborhood is inconsistent. In other words, a structural outlier has normal attributes while it may have several abnormal links. For example, those people from different communities but have a strong connection with each other can be regarded as structural outliers. As shown in Fig \ref{two-type}(a), 
there are two communities outlined with orange circles and structural outliers are those nodes with abnormal links to other communities. On the other hand, a contextual outlier has a consistent neighborhood structure while its attributes are corrupted, noisy, or significantly different from its neighbor nodes' attributes. For example, in Fig \ref{two-type}(b), suppose that the node in red is a football player while the nodes in green are music teachers. In this case, the node in red is regarded as a contextual outlier since it has a vast difference from its neighbors.  In the real world, datasets are much more complicated than this toy example and 
it is difficult to measure the degree of inconsistency among nodes.

\par
There have been various methods proposed to solve UNOD \cite{pygod}. They can be roughly divided into two categories, namely, non-deep  and deep-learning-based methods. Non-deep methods usually leverage traditional machine learning methods such as matrix factorization \cite{MF-1}, density-based clustering \cite{density-1}, and relational learning \cite{relation-1} to encode the graph information and detect outliers.  However, these methods fail to address the computational challenge with high-dimension data \cite{high-dimensional}. With the rapid prevalence of Graph Neural Networks (GNNs) \cite{GNN-survey}, more and more methods are based on deep learning \cite{DL-survey} and GNNs nowadays \cite{god-survey}. For example, DOMINANT \cite{GCNAE} employs two GNN autoencoders to reconstruct the attribute information and structure information.
According to the results reported in \cite{pygod}, most deep-learning-based methods have a much better performance than non-deep methods in detecting injected outliers. To unify the outlier injection process, \cite{pygod} adopts the outlier injection approach from \cite{GCNAE} as the standard injection approach.

\par
\begin{figure}[t]
		\centering
		\includegraphics[width=250.0 pt]{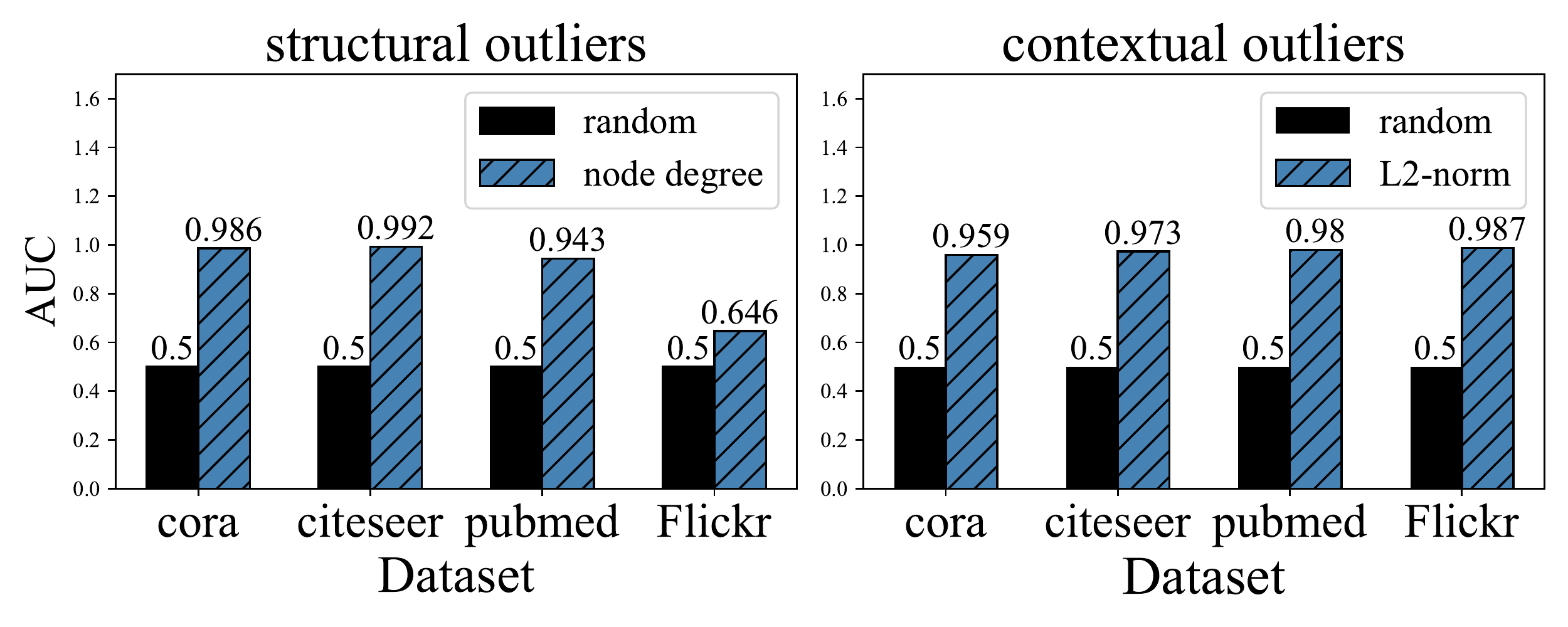}
  \vspace{-3mm}
		\caption{After injecting outliers in four datasets, node degree is employed to detect structural outliers while L2-norm is employed to detect contextual outliers. Both of them, compared to the random detector, achieve unexpectedly high scores. }%
		\label{data-leakage}
\end{figure}

 
\vspace{1.5mm}
\noindent \textbf{Challenge.} Although the recent deep-learning-based methods have achieved an excellent performance in UNOD, we find that the most widely-used outlier injection approach, \newtext{which is adopted by \cite{GCNAE,CoLA,AEGIS,AnomalyDAE,conad,pygod,interactive-gad,guide,Anemone,sl-gad,dagad}, will cause a serious data leakage issue.
Here, we refer to the data leakage \cite{data-leakage} in machine learning, which means the information strongly associated with the labels is leaked to the training dataset.}
After employing this approach to inject outliers, structural outliers will have a larger node degree than the average while attribute vectors of contextual outliers will have a larger L2-norm (a.k.a. Euclidean norm) than  expected. As a result, a simple solution only utilizing node degree or L2-norm of attribute vectors as the outlier score to detect the corresponding type of outliers can acquire a quite satisfying performance as shown in Fig \ref{data-leakage}. The metric of AUC \cite{auc} is adopted here to measure the detection performance. 
\newtext{Under such a serious data leakage issue in injected datasets, most existing algorithms cannot have a better performance than the simple solution. In addition, it is observed that existing algorithms have a performance drop in varied injection settings, in which the data leakage issue caused by the current injection approach is alleviated. 
Therefore, it is urgent to find out the cause of data leakage and reduce its impact. On the other hand, it is also necessary to exploit an effective UNOD algorithm, which has a superior performance and is robust to the data leakage issue.}
Moreover, the balance between structural and contextual outliers detection performance is little considered in existing works \cite{pygod}. An algorithm with unbalanced detection may only have  detection ability for a certain type of outliers. 
\newtext{It is found that existing algorithms focus more on contextual outliers than structural outliers when detecting them.}
To gain more feasible algorithms, comprehensive metrics for balance evaluation 
should be devised. 

\vspace{1.5mm}
\noindent \textbf{Our Solution.}
In this paper, we are devoted to analyzing the cause of data leakage and devising a superior outlier detection method for UNOD to achieve better performance in both the current injection setting and varied injection settings. 

In particular, we propose a novel variance-based model to detect  structural outliers, which adopts the variance of \newtext{representations} of neighbor nodes to detect structural outliers. To the best of our knowledge, it is the first time to employ the neighbor variance to detect outliers. 
\newtext{
Existing algorithms are based on either reconstruction of adjacency matrix 
 \cite{GCNAE,AnomalyDAE} or contrastive learning \cite{CoLA,sl-gad} to detect structural outliers. 
According to the definition, the  essence of a structural outlier is its inconsistent neighbors that come from different communities. However, existing algorithms cannot 
directly capture this essence. In this case, we devise a deep graph model to measure the inconsistency among neighbors by the variance of latent representations of neighbors, which captures the essence and gives a better utility for detection.} On top of this, a new framework, \underline{V}ariance-based \underline{G}raph \underline{O}utlier \underline{D}etection (VGOD), is also proposed to detect two types of outliers with a variance-based model and an attribute reconstruction model. To address the balance issue, we separately train two models to avoid  overtraining and normalize two types of outlier scores to eliminate the scale difference.
To evaluate the  detection balance on two outlier types, we introduce a new metric to measure the gap in the performance score. The experiments are conducted on 5 real-world datasets and the results demonstrate that VGOD achieves the best detection performance. 

\vspace{1.5mm}
\noindent \textbf{Contribution.} Our major contributions are as follows. 
\begin{enumerate}
    \item To the best of our knowledge, we are the first to identify the data leakage issue in the most widely-used outlier injection approach.
    \vspace{1mm}
    \item We analyze the cause of data leakage in depth, give suggestions for the future design of the outlier injection approach, \newtext{and purpose a new  approach for injecting structural outliers}.
    \vspace{1mm}
    \item We propose a novel variance-based model and a new VGOD framework, \newtext{which outperforms existing algorithms in detecting outliers} and alleviates the issue of the balance of detection.
    \vspace{1mm}
    \item Extensive experiments are conducted to demonstrate that our approach achieves the best detection performance and the detection balance.
\end{enumerate}


\section{Related Work}
\subsection{Graph Neural Network}

GNNs \cite{GNN-survey} are a group of neural network models which utilize the graph structure for network representation learning and various tasks. Among GNNs, GCN \cite{gcn} is one of the most influential models, which extends the convolutional operation in sequence or grid data to graph-structured data. Furthermore,  to aggregate messages from neighbors more flexibly, GAT \cite{gat} introduces an attention mechanism to learn the importance of each neighbor node. On the other hand, GraphSage \cite{graphsage} adopts the sampling-based method to aggregate the neighbor information  to work in large-scale graphs. From a topological learning perspective, GIN \cite{GIN} is a more expressive model than GCN and can achieve the same discriminative power as the 1-WL graph isomorphism test \cite{1-WT}. In our proposed framework, GNN plays a vital role in the network embedding representation of nodes. Generally, the GNN module in our framework can be set to any type of the above-mentioned GNNs.

\subsection{Unsupervised Node Outlier Detection on Attributed Networks}
UNOD on attributed networks has attracted considerable research interest in recent years due to its wide application in complex systems.  Radar \cite{Radar} utilizes the residuals of attribute information and its coherence graph structure for outlier detection. ANOMALOUS \cite{ANOMALOUS} conducts attribute selection and 
outlier detection jointly based on CUR decomposition and residual analysis. However, these methods have  computation limitations in high-dimension attributes  due to their shallow mechanisms. 

Quite a few studies based on the deep-learning technique have emerged recently \cite{god-survey}. Dominant \cite{GCNAE} builds deep autoencoders on top of GCN layers to reconstruct the adjacency and attribute matrices. AnomalyDAE \cite{AnomalyDAE} employs dual autoencoder architecture with cross-modality interactions and the attention mechanism to reconstruct the adjacency and attribute matrices. CoLA \cite{CoLA} and SL-GAD \cite{sl-gad} perform the UNOD task via contrastive self-supervised learning and random walk  to embed nodes. AEGIS \cite{AEGIS} studies UNOD in the inductive setting by utilizing generative adversarial ideas to generate potential outliers. DONE \cite{Done} employs deep unsupervised autoencoders to generate the network embedding which eliminates the effects of outliers at the same time. CONAD \cite{conad} adopts four data augmentation strategies and contrastive learning for outlier detection. GUIDE \cite{guide} replaces adjacency reconstruction with higher-order structure reconstruction to detect structural outliers. Under the manner of outlier injection, all these above deep methods show superior performance than non-deep methods in detecting these two types of outliers.  To evaluate UNOD algorithms, \cite{pygod} adopts the most widely-used outlier injection approach from \cite{GCNAE} as the standard injection method and provides unified benchmarks for UNOD, which facilitates  fairness for comparing different methods. 

Current UNOD methods have achieved an excellent performance. However, as demonstrated in Fig \ref{data-leakage}, the widely-used outlier injection approach exists a data leakage issue. To our surprise, simply using the combination of L2-norm and node degree to detect outliers can achieve state-of-the-art performance. Therefore, our work focuses on analyzing the cause of the data leakage issue  and designing a superior method. In addition, as mentioned in  \cite{pygod} that no current method has a balanced detection performance on two outlier types, we also consider the balance issue in our method.

\section{Preliminary}

In this section, we formally present some concepts which are used throughout this paper and define the problem. We use lowercase letters (e.g. $a$), bold lowercase letters (e.g. $\bm{x}$), uppercase letters (e.g. $X$), and calligraphic fonts (e.g. $\mathcal{V}$) to denote scalars, vectors, matrices, and sets, respectively.

\subsection{Graph Neural Network}

GNNs stack L layers of message-passing layers. Each layer performs a message passing through the given graph structure. After the initial node feature $ \bm{h_0} \in \mathbb{R}^{d_0} $ is transformed by L layers, the vector representation $ \bm{h_L} \in \mathbb{R}^{d_L} $ is learned for each node $ v $. Most message-passing layers can be expressed using the following rule:
\begin{equation}
    \bm{h_v^{(l)}} = \sigma ( \Psi^{(l)}(AGG(
    \{ \Phi^{(l)}(\bm{h_u^{(l-1)}})  , u \in \mathcal{N}_v \cup \{v\}  \}
    )) )
    \label{eq 1}
\end{equation}
where $\sigma(\cdot)$ is the active function, $ \Psi^{(l)}(\cdot)$ and $\Phi^{(l)}(\cdot)$ denote differentiable functions such as Multi-Layer Perceptrons (MLP). $ AGG(\cdot) $ denotes a differentiable, permutation invariant function (e.g. sum, mean, max) and $ \mathcal{N}_v$ denotes node $v$ 's direct linked neighbors.

Here, we introduce three commonly used GNNs, namely GCN, GAT, and GIN.

\textbf{Graph Convolution Network} (GCN) \cite{gcn} is the most widely-used GNN module, which adopts the propagation rule: 
\begin{equation}
    H^{(l+1)} = \sigma(\hat{A}H^{(l)}W^{(l)})
    \label{eq-gnn}
\end{equation}
where  $\hat{A}$ is the symmetric normalized adjacency matrix, $H^{(l)}$ is the $l^{th}$ hidden layer node representation, and $W^{(l)}$ is the parameters in the $l^{th}$ hidden layer.

\textbf{Graph Attention Network} (GAT) \cite{gat} flexibly aggregates messages from neighbors with calculated weight $ \alpha_{ij} $ (\newtext{vs. average weight adopted by GCN})  of each edge $\left \langle i,j \right \rangle$  as 
\begin{equation}
    \alpha_{ij} = \frac{exp(LeakyReLU(\bm{a}^T[W\bm{h_i}||W\bm{h_j}]))}{\sum_{k\in \mathcal{N}_i}exp(LeakyReLU(\bm{a}^T[W\bm{h_i}||W\bm{h_k}]))}
    \label{eq-gat}
\end{equation}
where $ \bm{a}  $ and $ W$ are the learnable weights. Layer mask ${(l)}$ is omitted for simplicity.

\textbf{Graph Isomorphism Network} (GIN) \cite{GIN} is the expressively more powerful GNN model, which follows the rule to propagate  messages as 
\begin{equation}
    H^{(l)} = \sigma( \Psi^{(l)}(A + (1+ \epsilon)\cdot I) H^{(l-1)})
\end{equation}
where $\epsilon$ can be a fixed or learnable scalar parameter, and $ I $ and $ A $ is the identity matrix and adjacency matrix, respectively.

\subsection{Unsupervised Node Outlier Detection on Attributed Networks}

\begin{definition}[Attributed Network]
  An attributed network can be denoted as $ \mathcal{G} = (\mathcal{V}, \mathcal{E},X) $,
  where $\mathcal V = \{v_1,v_2,...,v_n\}$ is the set of nodes ($ |\mathcal V| = n $), $\mathcal E$ is the set of edges ($ |\mathcal E| = m $), and $X \in \mathbb{R}^{n\times d} $ is the attribute matrix. The $i^{th}$ row vector $\bm{x_i}\in \mathbb{R}^d$ of the attribute matrix denotes the attribute information of the $i^{th}$ node. Node $i$'s direct  neighbors can be denoted as $\mathcal{N}_i $.
\end{definition}

With the aforementioned notations, the outlier detection problem on attributed network can be formally stated as a ranking problem.

\begin{definition}[Outlier Detection on Attributed networks]
  Given an attributed network $\mathcal G = (\mathcal V,\mathcal E, X) $, the goal is to learn an outlier score function $f(\cdot)$ to calculate the outlier score $o_i = f(v_i) $ for each node. The higher the outlier score $o_i$ is, the $i^{th}$ node is more likely to be a structural outlier or a contextual outlier.
  By ranking all the nodes with their outlier scores, the abnormal nodes can be detected according to their ranking.
  
\end{definition}

In this paper, we consider the setting of unsupervised node outlier detection (UNOD), which is generally adopted by  previous works. In this setting, none of the outlier labels of the nodes is given in the training phase.

\section{Data Leakage issue Analysis} 

As shown in Fig \ref{data-leakage}, the current widely-used outlier injection approach exists a serious data leakage issue. In this section, we analyze the data leakage issue in detail. For these two types of outliers, we first introduce the outlier injection approach from \cite{GCNAE}. Next, we theoretically analyze the cause of data leakage and give our suggestions for a better design of the outlier injection approach.

\subsection{Structural Outlier}
\subsubsection{Injection Approach} \label{str-inject-method}
The structural outliers are acquired by disturbing the topological structure of the graph. In a clique, nodes are fully connected. The intuition is that nodes in a clique should have a strong correlation with each other. Based on this, the outlier assumption is that a clique formed by unrelated nodes is structurally  abnormal.  The process of structural outlier injection is as followed. The first step is to specify the clique size $q$ and the number of cliques $p$. Next, for each clique, $q$ random nodes are chosen from the set of normal nodes and made fully connected as structural outliers. Therefore, total $p \times q$ structural outliers will be injected into the dataset. In previous works, the clique size $q$ is fixed to 15 for all datasets, and the value of $p$ is set according to the size of the dataset. 
\subsubsection{Cause Analysis}



It is obvious that the chosen structural outliers will have a higher node degree since additional edges are added to them. Table ~\ref{tab: data-for-od} shows that none of the three citation networks (Cora, Citeseer, PubMed) have an average node degree greater than 3. However, due to the above injection approach, all the outliers will have a node degree of at least more than 15. 




\subsection{Contextual Outlier}

\subsubsection{Injection Approach} \label{con-inject-method}
The contextual outliers are acquired by disturbing the attributes of nodes.  The injection process is as followed. Firstly, total $p \times q$ normal nodes will be chosen as contextual outliers, which have the same number as structural outliers. Next, for each chosen outlier node $v_i$, another $k$ nodes $\{ v_{c1}, v_{c2}, ..., v_{ck} \}$ are randomly sampled from $\mathcal{V}$ as a candidate set $\mathcal{V}_c$. For each $v_{ci}$ in $\mathcal{V}_c$, the Euclidean distance between the attribute vector $\bm{x_{ci}}$ of $v_{ci}$ and $\bm{x_i}$ of $v_i$ will be calculated. The attribute vector $\bm{x_{ci}}$ with the largest $\lVert \bm{x_{ci}} - \bm{x_i} \rVert_2$ will be used to replace $\bm{x_i}$ as the new attribute vector of $v_i$. The size of candidate set $k$ is set to 50, ensuring that the disturbance amplitude is large enough.
\vspace{0.1cm}
\subsubsection{Cause Analysis}
\newcommand{\norm}[1]{\lVert #1 \rVert}

To ensure a large enough disturbance of attributes, the above injection approach changes the $\bm{x_i}$ to $\bm{x_{ci}}$ with  the largest $\lVert \bm{x_{ci}} - \bm{x_i} \rVert_2$. However, such a strategy will lead to the L2-norm of the final chosen  $\bm{x_{ci}}$ (i.e. $\lVert \bm{x_{ci}} \rVert_2$) being more likely to be large. We make the following assumptions. 

\vspace{-0.2cm}
\begin{assumption}
Suppose both $ \bm{x_{ci}} \in \mathbb{R}^d$ and $\bm{x_i} \in \mathbb{R}^d$ are independently sampled from attribute matrix $X$. The rank of matrix $X$ is greater than $1$.
\label{assum:1}
\end{assumption}

\vspace{-0.3cm}

\begin{assumption}
For $ \bm{x_{ci}} \sim X$, $\bm{x_{ci}} = \lVert \bm{x_{ci}}\rVert_2 \cdot \Vec{e_{ci}}$, where $\lVert {\bm{x_{ci}}} \rVert_2$ and $\Vec{e_{ci}}$ are the modulo and direction of $\bm{x_{ci}}$, respectively. We assumed that $\lVert{\bm{x_{ci}}}\rVert_2$ and $\Vec{e_{ci}}$ are independently distributed.
\label{assum:independent}
\end{assumption}

\vspace{-0.2cm}
We use $P_r(x)$ to denote the possibility of $x$, then we have the following theorem. 
\begin{theorem}
$P_r(\norm{\bm{x_{ci}} - \bm{x_i}}_2 > 
\norm{\bm{x_{cj}} - \bm{x_i}}_2 
\Rightarrow
\norm{\bm{x_{ci}}}_2 > \norm{\bm{x_{cj}}}_2
)>0.5$
\label{theorem-1}
\end{theorem}

\vspace{-0.2cm}

\begin{proof}
We define $ D(\bm{x_{ci}},\bm{x_i}) = \norm{\bm{x_{ci}} - \bm{x_i}}_2$ . For notational convenience, we use $\bm{s}$ to refer to $\bm{x_{ci}}$ and $\bm{t}$ to refer to $\bm{x_i}$. Please note that both $\bm{s}$ and $\bm{t}$ are independently sampled from $X \in \mathbb{R}^{n \times d}$. 
\vspace{-2mm}
\begin{equation*}
\small
    \begin{split}
  D^2(\bm{s},\bm{t}) & = \sum_i^d(\bm{s}_i – \bm{t}_i)^2 \\
& = \sum_i^d \bm{s}_i^2-2\sum_i^d \bm{s}_i\bm{t}_i + \sum_i^d\bm{t}_i^2   \\
& = \norm{\bm{s}}_2^2 - 2\norm{\bm{s}}_2\norm{\bm{t}}_2 cos\alpha + \norm{\bm{t}}_2^2 \\ 
& = f(\norm{\bm{s}}_2)
\end{split}
\normalsize
\label{eq:theorem-proof}
\end{equation*}
where $\alpha$ is the angle between vector $\bm{s}$ and $\bm{t}$, and $\norm{\bm{s}}_2$ is the modulo of $\bm{s}$. From the above Equation,  we can regard $D^2(\bm{s}, \bm{t})$ as a quadratic function $f(\cdot)$ of $\norm{\bm{s}}_2$. Particularly,   $\norm{\bm{s}}_2 = \norm{\bm{t}}_2 cos\alpha$ is the symmetry axis for $f(\norm{\bm{s}}_2)$. According to the properties of a quadratic function in one variable, the function is monotonic on both sides of the symmetry axis. Therefore,
\begin{equation*}
\small
\begin{split}
    if \quad \norm{\bm{s}}_2 > \norm{\bm{t}}_2 cos\alpha \Rightarrow ( f(\norm{\bm{s}}_2) \uparrow  \quad \Rightarrow \norm{\bm{s}}_2 \uparrow) \\
    if \quad \norm{\bm{s}}_2 < \norm{\bm{t}}_2 cos\alpha \Rightarrow ( f(\norm{\bm{s}}_2) \uparrow  \quad \Rightarrow \norm{\bm{s}}_2 \downarrow)
\end{split}
\normalsize
\label{eq:monotonic}
\end{equation*}
where $\uparrow$  means increase and $\downarrow$  means decrease. In this case, we can draw the following conclusions.
\small
\begin{equation*}
\begin{split}
    P_r(\norm{\bm{s}}_2 > \norm{\bm{t}}_2 cos\alpha) & = P_r( f(\norm{\bm{s}}_2) \uparrow  \quad \Rightarrow \norm{\bm{s}}_2 \uparrow) \\
    & = P_r( D^2(\bm{s},\bm{t}) \uparrow  \quad \Rightarrow \norm{\bm{s}}_2 \uparrow) 
\end{split}
\label{eq:proof-possibility-core}
\end{equation*}
\begin{equation*}
    = P_r(\norm{\bm{x_{ci}} - \bm{x_i}}_2 > 
\norm{\bm{x_{cj}} - \bm{x_i}}_2 
\Rightarrow
\norm{\bm{x_{ci}}}_2 > \norm{\bm{x_{cj}}}_2
)
\label{eq:...}
\end{equation*}
\normalsize

Since $\bm{s}$ and $\bm{t}$ are independently sampled from attribute matrix $X$, we can draw 
\begin{equation*}
\vspace{-1mm}
    P_r(\norm{\bm{s}}_2 > \norm{\bm{t}}_2) = 0.5.
\end{equation*}

Due to the assumption that the rank of  $X$ is greater than $1$, the angle between $\bm{s}$ and $\bm{t}$ does not always equal  zero. Therefore, $P_r(cos\alpha \equiv 1)<1$. Note that $cos\alpha \leq 1$. Finally, we draw the following 
\begin{equation*}
P_r(\norm{\bm{s}}_2 > \norm{\bm{t}}_2cos\alpha) > 0.5
\end{equation*}
which means 
\small
\begin{equation*}
     P_r(\norm{\bm{x_{ci}} - \bm{x_i}}_2 > 
\norm{\bm{x_{cj}} - \bm{x_i}}_2 
\Rightarrow
\norm{\bm{x_{ci}}}_2 > \norm{\bm{x_{cj}}}_2
) > 0.5
\end{equation*}
\normalsize
\end{proof}

 Fig \ref{data-leakage} verifies our analysis that only utilizing the L2-norm of attribute vectors of nodes can achieve nearly 0.98 AUC score for all these four datasets when $k=50$. 

\begin{figure}[h]
		\centering
		\includegraphics[width=240.0 pt]{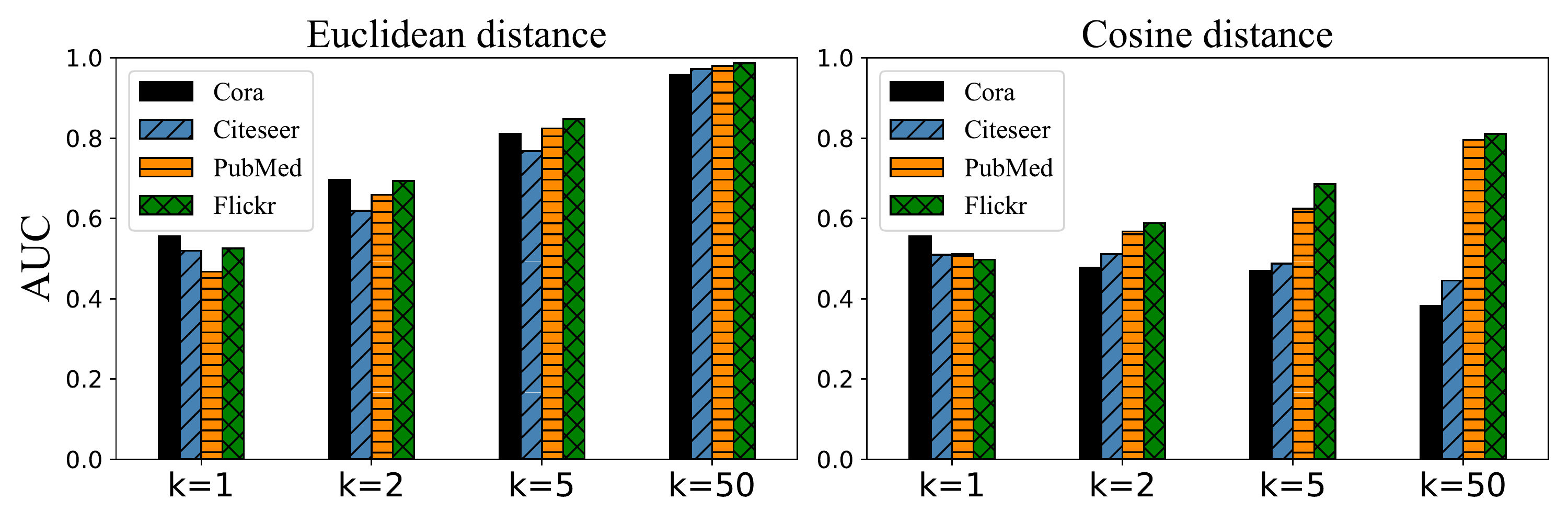}
		\caption{AUC  of L2-norm for contextual outliers injection with varying parameter of $k$ (size of candidate
set) and different distance measurements.}%
		\label{norm-change}
\end{figure}

\newtext{Further, we vary the parameter $k$ of the above injection approach.  As $k$ is set smaller, the data leakage issue is mitigated, which is shown in the left part of Fig \ref{norm-change}, indicating the large $k$ is the main cause for the serious data leakage issue. In the right part of Fig \ref{norm-change}, we also replace the Euclidean distance by cosine distance in the injection approach.  At this time, not all datasets have a data leakage issue when $k$ becomes large. Therefore, 
Euclidean distance is also a key factor for data leakage.}

\vspace{-0.3cm}

\subsection{Suggestion}
\newtext{Due to the serious data leakage issue, it is hard to figure out whether the outlier detection algorithm has an effect on detection or potentially exploits the leaked information of labels. According to our experiment in Section \ref{UNOD-exp}, a simple baseline only using the leaked information outperforms existing deep-learning-based solutions that need a long time to train. This can be explained from two aspects:}
\newtext{
\begin{enumerate}
    \item The data leakage issue caused by the current injection approach is too serious, which results in  little space for these algorithms to improve.
    \item Existing outlier detection algorithms are not effective enough.
\end{enumerate}}
\noindent \newtext{To mitigate data leakage caused by the current injection approach, we give the following suggestions on designing the new injection approach, better experiment, and parameter setting.}

\vspace{2mm}
\noindent \textbf{{\newtext{Suggestions for experiments and outlier injection.}}} \newtext{
We give the suggestion for  experiments and a new idea for outlier injection:}
\newtext{
\begin{itemize}
    \item Firstly, the data leakage issue should be examined. If it is hard to avoid  data leakage, then the leaked information should be compared to see the exact improvement.
    \item Secondly, some datasets contain  category labels for the node classification task. It is natural to think nodes with different labels come from different communities. Can we design a better injection approach based on this?
\end{itemize}}

 \vspace{2mm}
 \noindent \newtext{  
\textbf{Suggestions for structural outliers injection.}
The size of the structural outlier clique is much larger than the average node degree of graph, which leads to node degree being a signal for structural outliers. 
Therefore, we can set the injection clique size $q$ smaller for the current injection approach.} 

\newtext{On the other hand, in real world, higher node degrees are not a signal for structural outliers. For example, a famous person has many friends, but these friends are all in his or her community circle, so this person is not a structural outlier.
Therefore, to keep the distribution of node degree, replacing edges can be considered for a new injection approach.}

  \vspace{2mm} 
\noindent \newtext{ \textbf{Suggestions for contextual outliers injection.}
We have analyzed that the size of candidate set $k$ and Euclidean distance are two key factors to cause data leakage. Based on this, we give these suggestions:
\begin{itemize}
    \item Firstly, simply setting $k$ smaller can mitigate the data leakage for the current injection approach. 
    \item Secondly, replacing Euclidean distance with other distance measurements can be tried, such as cosine distance, shortest path distance, and so on. 
\end{itemize}} 

\subsection{Summary}
In summary, the current widely-used outlier injection approach will cause the data leakage issue, both in structural and contextual outlier injection. \newtext{We also give some suggestions for designing a new injection approach, better experiment, and parameter setting. We have applied some of our suggestions in our experiment to evaluate the effectiveness of existing outlier detection solutions and our solution.}

\section{methodology}

\begin{figure*}[!htbp]
\vspace{-2mm}
		\centering
		\includegraphics[width=490.0 pt]{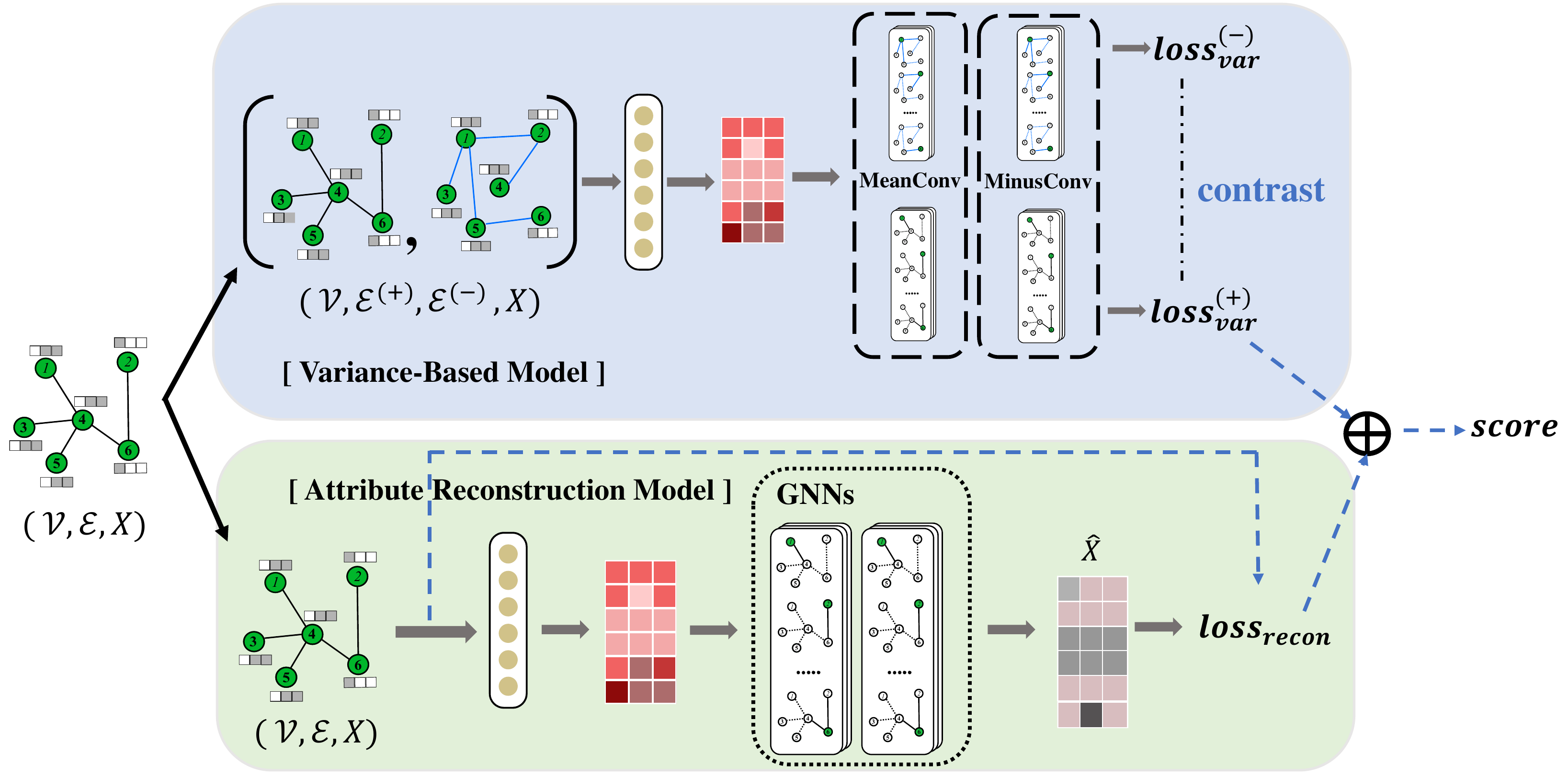}
		\caption{The overview of our proposed unsupervised node-level graph outlier detection framework VGOD. For a given attributed network $\mathcal{G}$, the variance-based model and attribute reconstruction model are employed to calculate the structural and contextual outlier score, respectively. The final score is the sum of two standardized scores. In the variance-based model (VBM), we use  a negative edge sampling technique to generate a corresponding negative edge set $\mathcal{E}^{(-)}$ per epoch which has the same number of edges as  $\mathcal{E}$. VBM is trained by the contrastive learning of $\mathcal{E}$ and $\mathcal{E}^{(-)}$.}%
		\label{VGOD-framework}
  \vspace{-1mm}
\end{figure*}

\begin{figure}[!htbp]
\vspace{-2mm}
		\centering
		\includegraphics[width=220.0 pt]{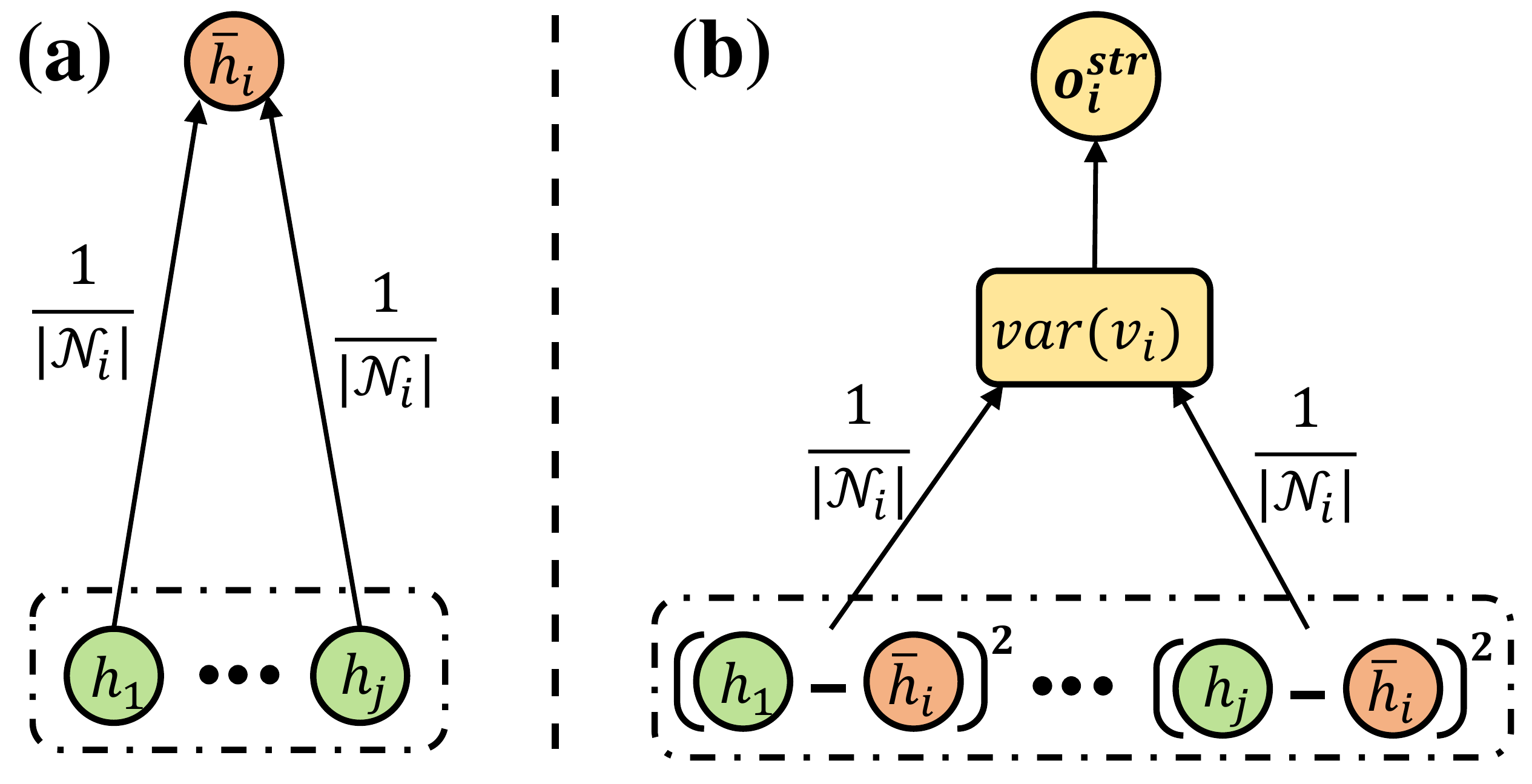}
		\caption{(a) The MeanConv Layer. (b) The MinusConv Layer. We calculate the variance of neighbor nodes’ low-dimension latent representation by (a) and (b).}%
		\label{two-Conv}
		 \vspace{-3mm}
\end{figure}

In this section, we are going to illustrate our proposed  framework VGOD in detail. Since current UNOD algorithms cannot outperform the simple baseline which only utilizes data leakage information, we propose our new framework VGOD, which combines a novel variance-based model and attribute reconstruction model. Specifically, the former model is for detecting structural outliers and the latter model is for detecting contextual outliers. Then we standardize the outlier scores outputted by two models and add them to get the final score. Fig \ref{VGOD-framework} presents  the whole architecture of VGOD framework.

\subsection{Variance-based Model}
In order to effectively detect  structural outliers, we propose a novel Variance-Based Model (VBM). 
To the best of our
knowledge, this is the first time to utilize the variance of neighbors to detect outliers.
Neighbor variance measures the consistency of neighbor nodes.
The bigger the variance, the less consistency it implies. In addition, our VBM has no bias on nodes with larger node degrees. 

\vspace{1mm}
\noindent{\textbf{Feature Transformation.}}
Before calculation of neighbor variance, we conduct the feature transformation $f_\theta(\cdot)$ for the original attribute matrix $X$ and get the low-dimension hidden representation matrix $H$ of nodes:
 \begin{equation}
    H = f_\theta(X)
    \label{struct-feature-trans} 
 \end{equation}
where $f_\theta(\cdot)$ denotes a neural network, such as $MLP(\cdot)$. The $i^{th}$ row vector $\bm{h_i}$  of the hidden representation matrix $H$ denotes the latent representation of the $i^{th}$ node.
In our experiment, we implement it with a linear transformation and L2-normalization  as :
 \begin{equation}
 \small 
 \begin{split}
     \hat{H} & = XW + \bm{b} \\
    \bm{h_i} & = \frac{\bm{\hat{h_i}}}{\norm{\bm{\hat{h_i}}}_2} 
 \end{split}
    \label{struct-feature-trans-W}  \normalsize
 \end{equation}
 where $W \in 
 \mathbb{R}^{d \times d_h} $ and $\bm{b} \in \mathbb{R}^{d_h} $ are the learnable parameters, $d$ and $d_h$ are the input dimension and hidden dimension of representation, respectively.
 \vspace{1mm}
 
\noindent{\textbf{Neighbor Variance.}}
In order to capture the consistency of neighbor nodes of a given node $v_i$, we calculate the variance of attribute vectors of neighbor nodes for $v_i$:
\begin{equation}
\small
    \bm{\overline{h}_i} = \frac{1}{|\mathcal{N}_i|} \sum_{j\in \mathcal{N}_i} \bm{h_j}
    \label{variance-mean}
\end{equation}
\vspace{-1mm}
\begin{equation}
    var(v_i) = \frac{1}{|\mathcal{N}_i|} \sum_{j\in \mathcal{N}_i}(\bm{h_j} -\bm{\overline{h}_i} )^2
    \label{variance-var}
\end{equation}
\vspace{-1mm}
\begin{equation}
    o^{str}_i = loss_{var}(v_i) = \lVert var(v_i) \rVert _1
    \label{variance-score}
\end{equation}\normalsize
where $\bm{\overline{h}_i}$ is the average of hidden representations of neighbor nodes of $v_i$. The L1-norm of $var(v_i)$  is applied as structural outlier score $o^{str}_i$ for node $v_i$. Since all components of the vector $var(v_i)$ are greater than 0, the L1 norm calculation is to simply sum components of the $var(v_i)$.
In order to efficiently calculate variance for each node, we implement the calculation of variance based on the message-passing scheme \cite{mpnn} and design two message-passing layers without parameters, namely MeanConv and MinusConv, as illustrated in Fig \ref{two-Conv}. Concretely, MeanConv is employed to calculate Eq. (\ref{variance-mean}) and MinusConv is used to calculate Eq. (\ref{variance-var}) as well as Eq. (\ref{variance-score}). 

\vspace{1mm}
\noindent{\textbf{Train.}}
In order to train VBM to learn the representation that a normal node has a low variance while a structural outlier has a high variance, the train objection for VBM can be formally defined as follows:
\begin{equation}
\small
    \mathop{min} \limits_{\theta} \mathbb{E}_{v_i \sim \mathcal{V}} [loss_{var}(v_i) - \frac{1}{|\mathcal{V}^{-\mathcal{N}_i}|} \sum_{j\notin \mathcal{N}_i} (\bm{h_j} - \frac{1}{|\mathcal{V}^{-\mathcal{N}_i}|}\sum_{u\notin \mathcal{N}_i } \bm{h_u})^2]
    \label{str-objective-1}
 \normalsize
\end{equation}
where $\mathcal{V}^{-\mathcal{N}_i} = \mathcal{V} \setminus \mathcal{N}_i$  is the non-neighbor node set of $v_i$.

We minimize the neighbor variance of a node while maximizing the variance of hidden representations of all the non-neighbor nodes. In this case, the model will avoid generating  the same hidden representations for all nodes. However, it is too expensive to maximize the variance of all non-neighbor nodes every time. In this case, we apply negative edge sampling each epoch to generate a network $\mathcal{G}^{(-)}$ whose edge set $\mathcal{E}^{(-)}$ has the same number of edges as $\mathcal{E}$. 
\newcommand{\edge}[1]{\left \langle #1 \right \rangle}
\begin{definition}[negative edge set]
    For a given attributed network $\mathcal{G} = \{\mathcal{V},\mathcal{E},X\}$,
    if $\mathcal{E}^{(-)}$ is the negative edge set of $\mathcal{G}$, then $if \edge{u,v} \in \mathcal{E}^{(-)} \Rightarrow \edge{u,v} \notin \mathcal{E}$ , $u,v \in \mathcal{V}$.
\end{definition}
\begin{definition}[negative network]
    For a given attibute network $\mathcal{G} = \{\mathcal{V},\mathcal{E},X\}$,
    we define the negative network $\mathcal{G}^{(-)} =\{\mathcal{V},\mathcal{E}^{(-)},X\} $, where $\mathcal{E}^{(-)}$ is the negative edge set of $\mathcal{G}$. 
\end{definition}

Therefore, we can utilize such a negative graph to maximize the variance of unrelated nodes. In other words, we randomly sample the same number of negative neighbors for each node $v_i$ and maximize  the neighbor variance calculated by these negative neighbors. Instead of maximizing the variance of all non-neighbor nodes, fewer nodes are required for computation by using negative sampling, which greatly saves time and space.

Therefore, for each node $v_i$, it has the ``related vs unrelated'' neighbor nodes pair, and corresponding $loss_{var}(v_i)^{(+)}$ and $loss_{var}(v_i)^{(-)}$ can be calculated respectively. The contrastive learning of neighbor nodes pair can be formalized as:
\begin{equation}
\small
\begin{split}
    loss_{var}^{(+)}(v_i) & = \lVert var(v_i,\mathcal{G}) \lVert_1  \\
    loss_{var}^{(-)}(v_i) & = \lVert var(v_i,\mathcal{G}^{(-)}) \lVert_1 \\
    loss^{str}(v_i) & = loss_{var}^{(+)}(v_i) - loss_{var}^{(-)}(v_i)
\end{split}
    \label{contrastive-learning}
     \vspace{1mm}
     \normalsize
\end{equation}
where $var(v_i,\mathcal{G})$ and $var(v_i,\mathcal{G}^{(-)})$ means we calculate the neighbor variance based on network $\mathcal{G}$ and $\mathcal{G}^{(-)}$, respectively.

Finally, we minimize the above $loss^{str}$ for all nodes in $\mathcal{V}$ as:
\begin{equation}
    \mathop{min} \limits_{\theta} \mathbb{E}_{v \sim \mathcal{V}} loss^{str}(v) 
\end{equation}

Thus, trained VBM can output a larger neighbor variance score for nodes with unrelated neighbor nodes and a relatively small score for nodes with related neighbor nodes. Consequently, we can utilize  VBM to detect  structural outliers.

\vspace{1.5mm}
 \noindent \newtext{\textbf{Can neighbor variance help detect contextual outliers?} We employ a simple technique, self-loop edge, to make neighbor variance have an effect on detecting contextual outliers besides structural outliers. In specific, self-loop edges are added to all nodes as
\begin{equation}
\vspace{-2mm}
\small
    \hat{\mathcal{N}}_i = \mathcal{N}_i \cup \{v_i\},  \forall v_i \in \mathcal{V}
    \label{self-loop}
\normalsize
\vspace{-0.5mm}
\end{equation}
where $\mathcal{N}_i$ is the neighbor set of node $v_i$.
As the attributes of a contextual outlier are significantly different from its neighbors, neighbor variance of it would be increased greatly after adding the self-loop neighbor when $|\mathcal{N}_i|$ (i.e. node degree of node $v_i$) is small.
This technique is optional and we employ it when the average node degree of graph is small. Our experiment in Section \ref{eff-self-loop} studies the effect of this technique.}

\subsection{Attribute Reconstruction Model}
We employ attribute reconstruction in the detection of contextual outliers.  Our attribute reconstruction Model (ARM) is  flexible that any popular GNN model can be used as the backbone to reconstruct the attributes of nodes.

\vspace{1mm}
\noindent{\textbf{Feature Transformation.}} Similar to VBM, we first transform the original attribute matrix $X$ to the low-dimension feature representation matrix $Z^{(0)}$ as:
\begin{equation}
\small
    \begin{split}
    \hat{Z} & = XW' + \bm{b'}\\        
    \bm{z^{(0)}_i} & = \frac{\bm{\hat{z}_i}}{\norm{\bm{\hat{z}_i}}_2}
    \end{split}
    \label{contextual-ft}
    \normalsize
\end{equation} 
where $W'$ and $\bm{b'}$ are the learning parameters, $\bm{z^{(0)}_i}$ is the $i^{th}$ row vector of $Z^{(0)}$.

\vspace{1mm}
\noindent{\textbf{GNN Layers.}} Then we employ $L$ GNN layers to transform $Z^{(0)}$ to $Z^{(L)}$ to fully absorb the message from neighbor nodes. The $l^{th}$ GNN Layer can be formalized as:
\begin{equation}
    Z^{(l)} = GNN^{(l)}(Z^{(l-1)},\mathcal{G})
    \label{GNN-operator}
\end{equation}
where $l^{th}$ operator $GNN^{(l)}(\cdot)$ can be implemented by any popular GNN model like GCN \cite{gcn}, GAT 
\cite{gat}, GIN \cite{GIN}, and so on.

\vspace{1mm}
\noindent{\textbf{Feature Retransformation.}}
Finally, we retransform the $Z^{(L)}$ to $ \hat{X}$, where $\hat{X} \in \mathbb{R}^{|\mathcal{V}| \times d}$ has the same shape as original attributes matrix $X$.
\begin{equation}
      \hat{X} = Z^{(L)} \hat{W} + \bm{\hat{b}}
  \label{context-recon}
\end{equation}
where $ \hat{X}$ is the reconstruction of the attribute matrix, $\hat{W}$ and $\bm{\hat{b}}$ are the weight and bias parameters.
Thus we can use the reconstruction attribute matrix $\hat{X}$ to calculate the reconstruction error, which is denoted as
\begin{equation}
\small
    o^{attr}_i = loss^{recon}(v_i) = \Vert \bm{\hat{x}_i} - \bm{x_i} \Vert ^2
    \label{recon_loss}
\end{equation}
\begin{equation}
     \mathop{min} \limits_{\theta} \mathbb{E}_{v \sim \mathcal{V}}
    loss^{recon}(v)
    \label{contextual-object}
\small
\end{equation}
By minimizing the above objective, trained ARM can detect contextual outliers.

\subsection{Outlier Detection}
As mentioned in \cite{pygod}, current UNOD algorithms fail to have balanced performance on two outlier types. 
The previous practice that combines contextual  and structural loss with a fixed weight during the training stage fails to balance the optimization of model parameters. Similarly, during the inference stage, combing the contextual  and structural score with a fixed weight  fails to achieve a balanced detection performance.
Therefore, we separately train our VBM and ARM with different epochs to avoid unbalanced optimization.

\vspace{1mm}
\noindent \textbf{Score combination.} After both VBM and ARM are well-trained, we employ the mean-std normalization on two types of outlier scores outputted by two models and sum scores to get the final score, which can be formalized as:
\begin{equation}
\small
    \begin{split}
    \widehat{o_i}^{str} & = \frac{o_i^{str} - \mu (\mathcal{O}^{str})}{std(\mathcal{O}^{str})} \\
    \widehat{o_i}^{attr} & = \frac{o_i^{attr} - \mu (\mathcal{O}^{attr})}{std(\mathcal{O}^{attr})} \\
    o_i & =  \widehat{o_i}^{str} +   \widehat{o_i}^{attr}
    \end{split}
    \label{mean-std-sum}
    \normalsize
\end{equation}
where $\mathcal{O}^{str}$ and $\mathcal{O}^{attr}$ denote the set of structural outlier scores and contextual outlier scores, respectively.
$\mu (\cdot)$ denotes the mean function,  
$std(\cdot)$ denotes standard deviation function.


By adopting Eq. (\ref{mean-std-sum}) as the final outlier score, our model can have a more balanced performance in detecting two types of outliers during the inference stage.
The overall procedure of our VGOD framework is described in Algorithm \ref{aig:1}. 

\begin{algorithm}
        \caption{The overall procedure of VGOD framework}
        \label{aig:1}
        \begin{algorithmic}[1] 
        \Require Attributed Network: $\mathcal{G} = (\mathcal{V},\mathcal{E},
        X)$, Training epochs for VBM: $Epoch_{VBM}$, Training epochs for ARM: $Epoch_{ARM}$
        \Ensure Well-trained VBM and ARM, outlier scores $\mathcal{O}$
        \State // \textit{Training phase}
         \For{ $i \in 1,2,...,Epoch_{VBM}$}     
            \State Generate the negative network $\mathcal{G}^{(-)} = (\mathcal{V},\mathcal{E}^{(-)},
        X)$ by negative edge sampling.
        \State Compute the neighbor variance of nodes in $\mathcal{G}^{(+)}$ and $\mathcal{G}^{(-)}$ via Eq. (\ref{struct-feature-trans-W})-(\ref{variance-score}).
        \State Update VBM with the loss function via Eq. (\ref{contrastive-learning}).
        \EndFor
         \For{ $i \in 1,2,...,Epoch_{ARM}$}

        \State Compute the reconstruction node attributes via Eq. (\ref{contextual-ft})-(\ref{context-recon}).

        \State Update ARM with the loss function via Eq. (\ref{recon_loss}).
        \EndFor
        \State // \textit{Inference phase}

        \State Compute the $\mathcal{O}^{str}$ and $\mathcal{O}^{attr}$ via VBM and ARM, respectively.
        \State Compute the final outlier scores $\mathcal{O}$ via Eq. (\ref{mean-std-sum}).

        \State \textbf{return} VBM, ARM, $\mathcal{O}$
        
        \end{algorithmic}
    \end{algorithm}

\subsection{Complexity Analysis}
 The complexity is mainly bounded by message-passing layers. For simplicity, the number of layers and the number of dimensions are considered  constant. The space and time complexity both are $O(|\mathcal{E}| + |\mathcal{V}|)$. 
 In addition, we are only using GNN layers and two linear layers to build our model. Since there are a large number of research on extending GNNs to larger networks, we can make use of various mini-batch training techniques such as \cite{graphsage,cluster-gcn,shadowKhop} to extend our model in a large-scale network without much effort.

\section{Experiment}
\label{Experiment-sec}

In this section, we conduct  experiments to illustrate the effectiveness of our proposed framework VGOD. Firstly, we describe the experiment settings including datasets, baselines, evaluation metrics, and computing infrastructures. Then, we conduct the Unsupervised Node Outlier Detection (UNOD) experiment to validate the effectiveness of our framework. \newtext{Next, we conduct structural outlier detection experiments under different injection parameters and a new injection approach, respectively.} Finally, we make  further analyses  for our approach, including efficiency and ablation study.  Additional experiments are conducted in \hyperlink{Appendix}{Appendix}. Our code is available at \url{https://github.com/goldenNormal/vgod-github}.

\subsection{Experiment settings}

\begin{table}[t]
\vspace{-2mm}
\centering
\caption{Datasets for UNOD Experiments.}
\label{tab: data-for-od}

\setlength{\tabcolsep}{1.5mm}{

\begin{tabular}{c|cccccc}
\hline
\textbf{Dataset} & \textbf{\#nodes} & \textbf{\#edges} & \textbf{\#attrs} & \textbf{\#avg\_deg} & \textbf{\#outliers}$^*$ & \textbf{\%outlier}$^*$ \\ \hline
Cora     & 2,706  & 5,429   & 1,433  & 2.01  & 150 & 5.5\%  \\
Citeseer & 3,327  & 4,732   & 3,703  & 1.42  & 150 & 4.5\%  \\
PubMed   & 19,717 & 44,338  & 500    & 2.25  & 600 & 3.0\%  \\
Flickr   & 7,575  & 239,738 & 12,407 & 31.65 & 450 & 5.9\%  \\
Weibo    & 8,405  & 407,963 & 64     & 48.5  & 868 & 10.3\% \\ \hline
\multicolumn{7}{l}{$^*$\#outlier and \%outlier are only for UNOD Experiments.}
\end{tabular}
}
\vspace{-3mm}
\end{table}

\subsubsection{Datasets}
We evaluate the proposed framework on five real-world datasets for UNOD on attributed networks, including four widely-used benchmark datasets with injected outliers and one dataset with labeled outliers.  These datasets, shown in Table \ref{tab: data-for-od}, include three citation networks\footnote{https://linqs.org/datasets/} (Cora, Citeseer, PubMed) and two social networks (Flickr\footnote{https://github.com/mengzaiqiao/CAN} and Weibo\footnote{https://github.com/zhao-tong/Graph-Anomaly-Loss/tree/master/data}). Only the Weibo dataset contains labeled outliers.

\subsubsection{Baselines}
\begin{table*}[!h]
\vspace{-2mm}
\centering
\caption{The Comparison of Baselines.}
\label{tab:baseline}
\setlength{\tabcolsep}{4mm}{
\begin{tabular}{|c|c|c|c|c|c|}
\hline
Baseline   & Time Complexity  & Contrastive Learning & Reconstruction & Score Combination & Inductive Inference \\ \hline
Dominant \cite{GCNAE}   & $O(|\mathcal{E}| + |\mathcal{V}|^2)$           &   $\times$              & \checkmark               & \checkmark         & \checkmark       \\
AnamalyDAE \cite{AnomalyDAE} & $O(|\mathcal{E}| + |\mathcal{V}|^2)$            & $\times$                    & \checkmark               & \checkmark       &   $\times$              \\
DONE \cite{Done}      & $O(|\mathcal{V}|K)^1$            & $\times$                    & \checkmark               & \checkmark            & \checkmark      \\
CoLA \cite{CoLA}      & $O(c|\mathcal{V}|R(c+\delta))^2$      & \checkmark       & $\times$      & $\times$ & \checkmark   \\
CONAD \cite{conad}     & $O(|\mathcal{E}| + |\mathcal{V}|^2)$            & \checkmark                    & \checkmark               & \checkmark          & \checkmark        \\
VGOD (Ours)       & $O(|\mathcal{E}| + |\mathcal{V}|)$              & \checkmark                    & \checkmark               & \checkmark & \checkmark     \\ \hline         
\multicolumn{6}{l}{ $^1$ $K$ denotes the number of sampling neighbors for each node}
\\
\multicolumn{6}{l}{$^2$ $R$ denotes the number of sampling rounds, $c$ denotes the number of nodes within the local subgraph, and $\delta$ denotes the average degree of the network.}
\\
\end{tabular}
}
\vspace{-3mm}
\end{table*}

We compare our proposed  framework VGOD with the recent five deep-learning-based SOTA models. 
These baselines are summarized in Table \ref{tab:baseline}. Column time complexity indicates the time complexity of inference. For simplicity, the number of layers and the number of dimensions are considered as constant. If the model outputs more than one score for outliers like VGOD, then we consider that it has the feature of score combination. 
\newtext{If the hyperparameters of the model (e.g., the number of layer units) are not coupled  to the  number of nodes or edges of the training graph, then we regard it can perform inductive inference, which means a trained model can be directly used for detecting outliers on a new graph with the same attribute schema.}
\newtext{Since none of the outlier labels  is given in the training phase, our experiments are conducted in the transductive setting, which is consistent with existing unsupervised outlier detection works.}

Due to data leakage in the outlier injection approach, we also design a simple baseline for comparison and evaluation, which only utilizes the leaked information (i.e. node degree and L2-norm of attribute vectors).
We name it DegNorm. 

DegNorm
    adopts node degree as structural outlier score while L2-norm of attribute vectors of nodes are adopted as contextual outlier score. The mean-std normalization is applied to two scores. The final outlier score is the sum of these two scores which have been normalized. The calculation of $o_i^{str}$ and $o_i^{attr}$ can be formalized as:
    \begin{equation}
        \begin{split}
         o_i^{str} \quad   & = |\mathcal{N}_i| \\
         o_i^{attr} & = \norm{\bm{x_i}}_2 \\
        \end{split}
        \label{eq:degnorm}
    \end{equation}
    where $\mathcal{N}_i$ is the neighbor node set of node $v_i$, $\bm{x_i}$ is the attribute vector of node $v_i$.

\subsubsection{Evaluation Metrics}

We use \textit{Area Under receiver operating characteristic Curve} (AUC) to measure. In specific, AUC evaluates the degree of alignment between the
outlier score and the ground truth label under varying thresholds:
\begin{equation}
\small
\label{auc}
   AUC = \frac{1}{| \mathcal V^{+} || \mathcal V^{-} | } \sum_{v^{+}_i\in \mathcal V^{+}} \sum_{v^{-}_j\in \mathcal V^{-}}    (\mathbb{I}(f(v^{+}_i) < f(v^{-}_j)))
\normalsize
\end{equation}
where $\mathcal{V}$, $\mathcal V^{-}$, and $\mathcal V^{+}= \mathcal{V} \setminus \mathcal V^{-}$  are the set of all nodes, the set of all outlier nodes, and the set of all normal nodes respectively, $\mathbb{I}(\cdot)$ is the indicator function and $f(v_i)$ is the outlier score of node $v_i$ given by one outlier detector.  \newtext{To explore the utility of the model for different outliers, we extend the concept of AUC.
Generally, $AUC(\mathcal{V}_L,\mathcal{O})$ means using $\mathcal{V}_L$ as the set of outliers to be detected, $\mathcal{O}$ as the outlier scores to calculate the AUC. In other words, $\mathcal{V}_L$ defines the outlier labels.
Particularly, $AUC = AUC(\mathcal V^{-},\mathcal{O})$.}
\newtext{In addition, if  a model can output the structural and contextual outlier scores  like our VGOD, then $AUC(\mathcal V^{-},\mathcal{O}^{str})$ and $AUC(\mathcal V^{-},\mathcal{O}^{attr})$ can be calculated.}

We also propose $AucGap$ to evaluate the balanced detection performance for different types of outliers, which can be formalized as below:
\begin{equation}
\small
     AucGap  = max \{
     \frac{AUC(\mathcal V^{str},\mathcal{O})}{ AUC(\mathcal V^{attr},\mathcal{O})},\frac{ AUC(\mathcal V^{attr},\mathcal{O})}{AUC(\mathcal V^{str},\mathcal{O})}
     \}
\label{eq:AucGap}
\normalsize
\end{equation}
where $\mathcal{V}^{str}$ and $\mathcal{V}^{attr}$ are structural outliers set and contextual outliers set, respectively. $AucGap$ aims to calculate the gap between  the model’s AUC score for two types of outliers. The lower the $AucGap$ is, the more balanced  detection performance it indicates.

    
    

\subsubsection{Computing Infrastructures}
Our proposed learning framework is implemented using PyTorch 1.11.1 and PyTorch Geometric 2.1.0. All experiments are conducted on a computer with Ubuntu 16.04 OS, i7-9750H CPU, and a Tesla V100 (32GB memory) GPU.

\subsection{Unsupervised Node Outlier Detection}
\label{UNOD-exp}
We first conduct the UNOD experiment to verify the effectiveness of our proposed framework. UNOD experiment hereinafter refers to this experiment.


\begin{table*}[htbp]
  \centering
  \caption{\newtext{AucGap of UNOD experiment.}}
  \label{tab: gap-result-od}
   \vspace{-2mm}
\renewcommand{\arraystretch}{1.1} 
    \begin{tabular}{c|ccc|ccc|ccc|ccc}
    \toprule
    \multirow{2}[2]{*}{Model} & \multicolumn{3}{c|}{Cora} & \multicolumn{3}{c|}{Citeseer} & \multicolumn{3}{c|}{PubMed} & \multicolumn{3}{c}{Flickr} \\
          & \textit{AucGap} &  str  & context & \textit{AucGap} & str   & context & \textit{AucGap} & str   & context & \textit{AucGap} & str   & context \\
    \midrule
    Dominant & 1.312  & \textit{\textcolor[rgb]{ 1,  0,  0}{0.696}} & 0.913  & 1.165  & 0.755  & 0.880  & 1.652  & \textit{\textcolor[rgb]{ 1,  0,  0}{0.600}} & 0.990  & 2.029  & \textit{\textcolor[rgb]{ 1,  0,  0}{0.486}} & 0.986  \\
    AnomalyDAE & 1.161  & 0.895  & 0.771  & 1.070  & 0.864  & 0.808  & 1.118  & 0.933  & 0.834  & 1.860  & \textit{\textcolor[rgb]{ 1,  0,  0}{0.521}} & 0.969  \\
    DONE  & 1.217  & 0.922  & 0.758  & \textbf{1.016 } & 0.872  & 0.886  & 1.217  & 0.836  & 0.687  & 1.557  & \textit{\textcolor[rgb]{ 1,  0,  0}{0.578}} & 0.900  \\
    CoLA  & \uline{1.127}  & 0.943  & 0.837  & 1.188  & 0.953  & 0.802  & \uline{1.054}  & 0.954  & 0.905  & \uline{1.395}  & \textit{\textcolor[rgb]{ 1,  0,  0}{0.622}} & 0.868  \\
    CONAD & 1.877  & \textit{\textcolor[rgb]{ 1,  0,  0}{0.513}} & 0.964  & 2.236  & \textit{\textcolor[rgb]{ 1,  0,  0}{0.434}} & 0.972  & 2.417  & \textit{\textcolor[rgb]{ 1,  0,  0}{0.404}} & 0.976  & 2.066  & \textit{\textcolor[rgb]{ 1,  0,  0}{0.478}} & 0.987  \\
    \midrule
    DegNorm & 1.132  & 0.936  & 0.827  & 1.116  & 0.979  & 0.877  & 1.093  & 0.861  & 0.941  & 1.822  & \textit{\textcolor[rgb]{ 1,  0,  0}{0.527}} & 0.960  \\
    VGOD  &\textbf{1.072}& 0.970  & 0.905  & \uline{1.026}  & 0.986  & 0.961  & \textbf{1.021} & 0.962  & 0.983  & \textbf{1.066} & 0.838  & 0.893  \\
    \bottomrule
    \end{tabular}%
    \vspace{-1mm}
  \label{tab:addlabel}%
\end{table*}%

\subsubsection{Injection Setting}
\label{leak-inject-setting}
We adopt the most widely-used outlier injection approach as mentioned in Section \ref{str-inject-method} and Section \ref{con-inject-method}. \newtext{ We keep the same injection parameter setting with \cite{CoLA,GCNAE,guide,interactive-gad,sl-gad,Anemone} to have a fair comparison (i.e., $q = 15$, $k=50$ for all datasets and $p= 5,5,20,15$ for Cora, Citeseer, PubMed, Flickr, respectively).}
The statistics of these  datasets are demonstrated in Table \ref{tab: data-for-od}. Only Weibo contains the labeled outliers while other datasets contain injected outliers. Note that $AucGap$ can only be calculated on these injected datasets.

\subsubsection{Parameter Setting}
For each algorithm, we run 5 times and calculate the average score to list here. For our proposed framework VGOD, we fix the embedding dimension to 128 for both the variance-based model (VBM) and attribute reconstruction model (ARM). We set the learning rate to 0.005 for all injected datasets and 0.01 for Weibo. Two layers of GAT are adopted as the GNN module in ARM  and the row-normalization to the attribute vectors is applied in Weibo. \newtext{We employ self-loop edge technique in Cora, Citeseer, PubMed, and Weibo for VGOD.}
We directly run the code in \cite{CoLA} to inject outliers. For all baselines, we adopt the default parameter setting in their code except the number of training epochs. We stop training their model as long as their AUC score reaches its peak.  In this case, the performance can be promised to be better or equal to the performance of their default parameter setting. 
For our approach, we train ARM 100 epochs and VBM 10 epochs for all datasets since it has already significantly outperformed  baselines in a fixed number of training epochs. In fact,  our two models require fewer epochs to converge. Adam optimizer is employed to train our models. We adopt the AUC score  of Weibo published in \cite{pygod} for the baseline Dominant, AnomalyDAE, DONE, and CONAD.

\begin{table}[!t]
\caption{AUC for UNOD experiment.}
\label{tab: auc-result-od}
\centering
\vspace{-2mm}
\begin{tabular}{c|cccc|c}
\hline
{Model}    & {Cora}   & {Citeseer} & {PubMed} & {Flickr} & {Weibo}  \\ \hline
{Dominant
}      & 0.8134          & 0.8250            & 0.7999          & 0.7440          & 0.925$^*$         \\
{AnomalyDAE} & 0.8433          & 0.8441            & 0.8898          & 0.7524          & \underline{ 0.928}$^*$   \\
{DONE}     & 0.8498          & 0.8800            & 0.7664          & 0.7482          & 0.887$^*$         \\
{CoLA}     & 0.8790          & 0.8861            & \underline{ 0.9214}    & \underline{ 0.7530}    & 0.748         \\
{CONAD}    & 0.7456 
        & 0.7078 
          & 0.6930 
        & 0.7395 
        & 0.927$^*$        \\ \hline
{DegNorm} & \underline{ 0.8928}    & \underline{ 0.9385}      & 0.9074          & 0.7515          & 0.893          \\
{VGOD}     &\textbf{0.9503}& \textbf{0.9845}& \textbf{0.9813} & \textbf{0.8773} & \textbf{0.9765} \\ \hline
\multicolumn{6}{l}{ $^*$ denotes the result reported in \cite{pygod}}
\end{tabular}
\vspace{-4mm}
\end{table}

\subsubsection{Result Analysis}
The AUC scores and $AucGap$  scores are shown in Table \ref{tab: auc-result-od} and Table \ref{tab: gap-result-od}. \newtext{$AUC(V^{str},\mathcal{O})$ and $AUC(V^{attr},\mathcal{O})$ are listed in the column of \textit{str} and \textit{context}.} The best score is in bold while the second best is underlined. \newtext{The AUC scores lower than 0.7 are red and italicized in Table \ref{tab: gap-result-od}.} According to the results, we have the following observations:
\begin{itemize}
    \item Our proposed framework VGOD achieves the highest AUC score for all datasets while achieving the overall highest $AucGap$ among all datasets. There are several reasons for such performance. Firstly, our variance-based model significantly improves the ability to detect structural outliers.
    Secondly, we separately train the model to prevent each component from being over-trained. Thirdly, we adopt mean-std normalization  to eliminate the scale difference between the two scores which gives a more balanced detection performance.

    \item DegNorm also achieves SOTA performance compared to other baselines.
    
    \item \newtext{In Table \ref{tab: gap-result-od}, it is observed that all baselines can not have a good detection performance on structural outliers in the Flickr dataset and achieve a poorly balanced detection.}
    
\end{itemize}
 \newtext{Though the $AucGap$ of VGOD in Citeseer is slightly lower than DONE,  its detection performance is already balanced. Moreover, both $AUC(V^{str},\mathcal{O})$ and $AUC(V^{attr},\mathcal{O})$ of VGOD are much higher than that of DONE.}

\subsection{Structural Outlier Detection under different injection parameters}
\label{Structural-detection-parameters}
Further, we conduct the structural outlier detection experiment with varied injection parameters to explore the effectiveness of our variance-based model (VBM) in depth.

\subsubsection{Injection Setting}
\begin{figure*}[!htbp]
\vspace{-2mm}
		\centering
		\includegraphics[width=500.pt]{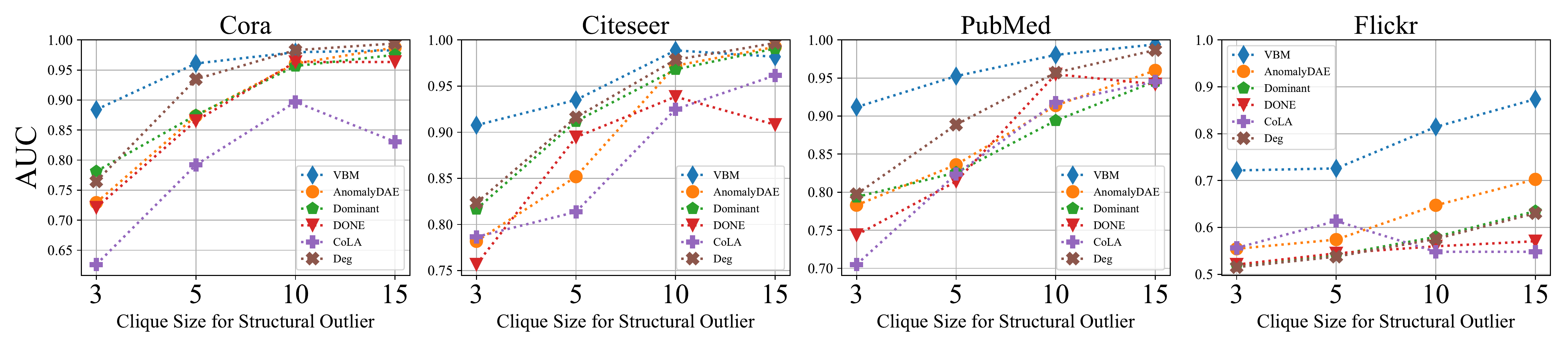}
  \vspace{-2mm}
		\caption{Comparison of the detection performance with the varying  clique size parameter. Each polyline represents the AUC values of a model on several groups of structural outliers with different clique sizes.}%
		\label{fig:all_str_auc}
  \vspace{-2mm}
\end{figure*}

We vary the parameter $q$ of injected clique size of structural outliers to $Q = \{3, 5, 10, 15\}$. For each dataset $\mathcal{D}_i$, we inject 4 groups of structural outliers $\{ \mathcal{V}^{q=3},\mathcal{V}^{q=5},\mathcal{V}^{q=10},\mathcal{V}^{q=15} \}$ into $\mathcal{D}_i$. Each group has the same number of outliers, which is set to 2\% of the total number of nodes, i.e. $|\mathcal{V}^{q=
Q_i}| = 2\% \cdot |\mathcal{V}|$. The outlier set $\mathcal V^{-}$ is the union of 4 groups of structural outliers set. We report the \newtext{$AUC(\mathcal V^-,\mathcal{O}^{str})$}  in Table \ref{tab: result-vbm} and the AUC score of each group \newtext{$AUC(V^{q=Q_i},\mathcal{O}^{str})$} is shown in Fig \ref{fig:all_str_auc}. 
\newtext{Note that $\mathcal{O}^{str}$ of VGOD is the output of VBM.
The injected outliers are all structural outliers.}

\subsubsection{Parameter Setting}
We keep the same parameter setting for VBM and other baselines as the UNOD experiment except that we train baselines and VBM until their AUC scores 
 reach the peak. Since we fail to get a reasonable result for CONAD, we do not list the result of it. We also evaluate the performance of simple baseline \textbf{Deg}, which only utilizes the node degree as an outlier score for comparison. 
For all other baselines, if their model outputs multiple scores (e.g., $o_i$, $o_i^{str}$, $o_i^{attr}$), we adopt the score with the highest AUC as its structural score.

\begin{table}[htbp]
\vspace{-2mm}

\caption{Auc for structural outlier detection under different injection parameters}
\label{tab: result-vbm}
\centering
\begin{tabular}{l|llll}
\hline
{Model}      & {Cora}   & {Citeser} & {PubMed} & {Flickr} \\ \hline
{Dominant}   & 0.9227          & 0.9467           & 0.8878          & {\ul 0.5715}    \\
{AnomalyDAE} & 0.9127          & 0.9219           & 0.8968          & 0.6253          \\
{DONE}       & 0.9034          & 0.8985           & 0.8868          & 0.5516          \\
{CoLA}       & 0.8073          & 0.8919           & 0.8698          & 0.5712          \\ \hline
{Deg}        & {\ul 0.9467}    & {\ul 0.9541}     & {\ul 0.9333}    & 0.5671          \\
{VBM}        & \textbf{0.9815} & \textbf{0.9816}  & \textbf{0.9893} & \textbf{0.8003} \\ \hline
\end{tabular}
\vspace{-1mm}
\end{table}

\subsubsection{Result Analysis}
According to results in Table \ref{tab: result-vbm} and Fig \ref{fig:all_str_auc}, we have the following observations: 
\begin{itemize}
    \item VBM achieves the best AUC score for all datasets in Table \ref{tab: result-vbm}. In addition, VBM has a huge performance gain in  Flickr.

    \item As shown in Fig \ref{fig:all_str_auc}, when the clique size is reduced,  the performance of VBM declines the least compared to other baselines. Therefore, the performance of VBM is the most robust to varied injection settings.
    
    \item Deg that directly utilizes a node's degree outperforms  other baselines in Cora, Citeseer, and PubMed. 
\end{itemize}

\subsection{Structural Outlier Detection under a new injection approach}
\newtext{
In this subsection, we design a new approach to inject structural outliers without data leakage. We conduct the following experiment for evaluation. }

\subsubsection{Injection approach}
\newtext{
Since  all these datasets have  category labels for the node classification  task, it is natural to think that nodes with different labels are  from different communities. In our opinion, structural outliers do not necessarily form  clusters. We generate structural outliers by replacing their original neighbors (both in and out) with nodes uniformly sampled from other communities.
In this manner, the degree distribution of all outliers is not changed.
The number of outliers is set as 10\% of the number of nodes. The injected outliers are all structural outliers. }

\begin{table}[htbp]
\vspace{-2mm}
  \centering
  \caption{Auc for structural outlier detection under a new  injection approach}
  \label{tab:novel-str-auc}
\renewcommand{\arraystretch}{1.1} 
    \begin{tabular}{c|cccc}
    \toprule
          & {Cora} & {Citeseer} & {PubMed} & {Flickr} \\
    \midrule
    {Dominant} & \uline{0.838}  & \uline{0.770}  & \uline{0.853}  & \uline{0.917}  \\
    {AnomalyDAE} & 0.770  & 0.673  & 0.566  & 0.898  \\
    {DONE} & 0.762  & 0.664  & 0.659  & 0.541  \\
    {CoLA} & 0.658  & 0.743  & 0.752  & 0.632  \\
    {CONAD} & 0.793  & 0.770  & 0.779  & 0.495  \\
    \midrule
    {VBM} & \textbf{0.935 } & \textbf{0.907 } & \textbf{0.858 } & \textbf{0.958 } \\
    \bottomrule
    \end{tabular}%
  \label{tab:addlabel}%
\end{table}%

\subsubsection{Result Analysis}

\newtext{
We keep the same parameter setting for VBM and all baselines  as  the experiment in Section \ref{Structural-detection-parameters}.
Table \ref{tab:novel-str-auc} lists the $AUC(\mathcal{V}^-,\mathcal{O}^{str})$.  Our VBM is still the most effective model, which outperforms others with a significant gap. 
This further verifies the effectiveness of neighbor variance to detect structural outliers.
}

\subsection{Further Analysis}
In this subsection, we make further analyses of our proposed framework. 

\subsubsection{Efficiency of model inference}
\begin{figure}[!htbp]
\vspace{-2mm}
		\centering
		\includegraphics[width=150.pt]{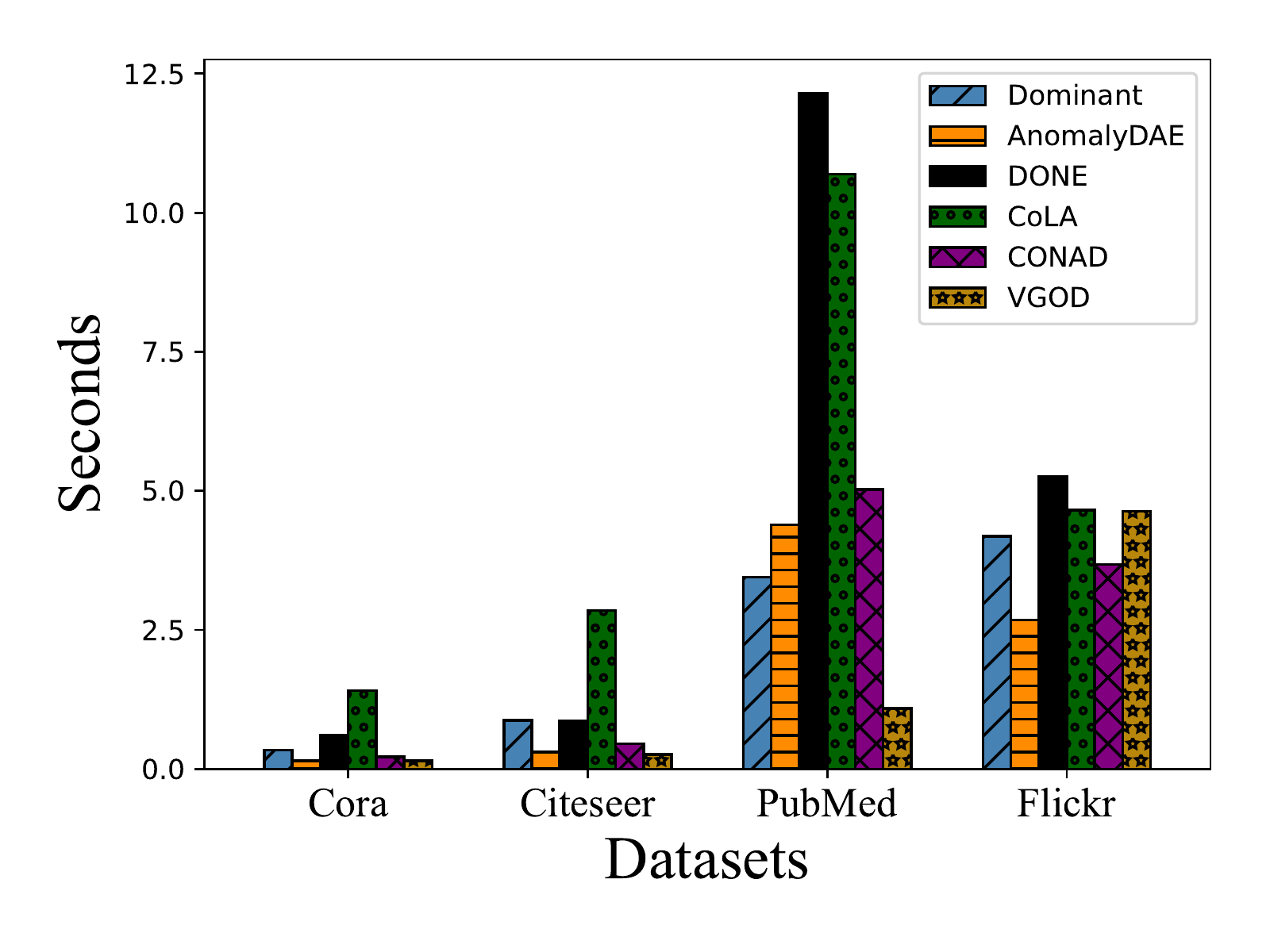}
		\vspace{-3mm}
		\caption{Training time per epoch in seconds.}%
		\label{fig:eff}
		\vspace{-1mm}
\end{figure}

We calculate the time for each model to use the CPU for training and inference at the setting of the UNOD experiment.
The training time per epoch of all models (in seconds) is shown in Fig \ref{fig:eff}.
 In Table \ref{tab: inference time123}, we list the inference time in seconds. The inference time of the model is roughly the same as the training time per epoch, except for CoLA. 
For all datasets, our VGOD framework completes inference in a relatively short time. For datasets with a large number of nodes, such as PubMed, our model takes significantly less time than other models due to the linear relationship to the number of nodes. Since CoLA requires multiple rounds of sampling for inference, its computational cost is much higher than other models.

\begin{table}[!ht]
\vspace{-2mm}
\caption{inference time of models (in seconds).}
\label{tab: inference time123}
\centering
\begin{tabular}{c|cccc}
\hline
\multicolumn{1}{c|}{{Model}} & {Cora} & {Citeseer} & {PubMed} & {Flickr} \\ \hline
{Dominant}   & 0.102          & 0.235          & 3.021        & 4.183          \\
{AnomalyDAE} & 0.147          & 0.303           & 4.390          & { 2.493}    \\
{DONE}       & 0.604          & 0.865           & 12.147          & 5.256           \\
{CoLA}       & 413        & 752        & 3266       & 910        \\
{CONAD}      & { 0.093}    & { 0.201}    & { 2.823}    & \textbf{1.379} \\ \hline
{VGOD}       & \textbf{0.088} & \textbf{0.145} & \textbf{0.874} & 3.899     \\ \hline    
\end{tabular}
\vspace{-2mm}
\end{table}

\begin{figure*}[ht]
\vspace{-2mm}
		\centering
		\includegraphics[width=500.pt]{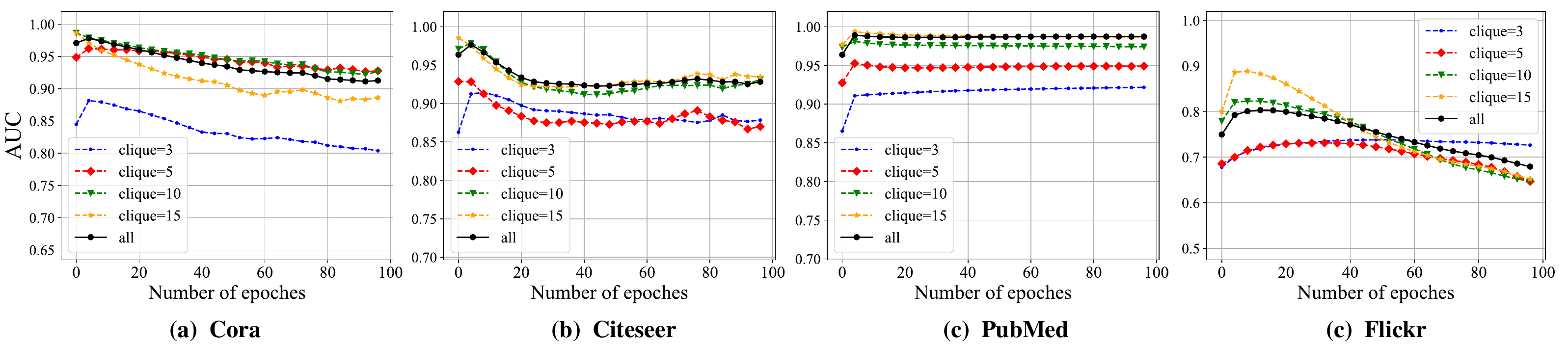}
  \vspace{-3mm}
		\caption{AUC variation trend of the variance-based model during the training of VBM. Each polyline represents a group of structural outliers with a certain clique size.}%
		\label{fig:vbm-ablation-epoch}
  \vspace{-2mm}
\end{figure*}

\subsubsection{Effect of the number of epochs for VBM}
We investigate the AUC variation trend of VBM during training. As shown in Fig \ref{fig:vbm-ablation-epoch}, VBM shows a high AUC score at the beginning, and the AUC score reaches the peak after only a few epochs of training. Afterward, as the training progresses, the AUC score slowly decreases due to  overfitting. Different group of structural outliers shows a similar trend while the group of smaller clique size shows a later overfitting time point.

\subsubsection{Effect of different GNN Layers for ARM}
 We investigate the effect of different GNN layers in ARM. We replace different GNN layers in the UNOD experiment for research. Table \ref{tab: gnn-ablation-auc} and Table \ref{tab: gnn-ablation-gap} show the AUC and $AucGap$ respectively  on four datasets. It is observed that GAT outperforms other GNNs significantly on the Weibo. For other datasets, their AUC and $AucGap$ scores are comparable.

\begin{table}[!htbp]
\vspace{-2mm}
\caption{AUC values comparison for different GNN layers.}
\label{tab: gnn-ablation-auc}
\centering
\setlength{\tabcolsep}{1.4mm}{
\begin{tabular}{c|ccccc}
\hline
{Model}      & {Cora} &{Citeseer} & {PubMed}                     & {Flickr} & {weibo}  \\ \hline
{VGOD (GIN)} & 0.9503          & 0.9845 & 0.9801 & 0.8773 & 0.9093 \\
{VGOD (GCN)} & \textbf{0.9566} & 0.9867 & 0.9802 & 0.8735 & 0.9154 \\
{VGOD (GAT)} & 0.9560        & \textbf{0.9868}   &{\textbf{0.9813}}& \textbf{0.8835} & \textbf{0.9765} \\ \hline
\end{tabular}
}
\vspace{-2mm}
\end{table}

\begin{table}[!htbp]
\vspace{-2mm}
\caption{AUCGAP values comparison for different GNN layers.}
\label{tab: gnn-ablation-gap}
\centering
\begin{tabular}{c|cccc}
\hline
{Model}      & {Cora}   & {Citeseer} & {PubMed}                     & {Flickr} \\ \hline
{VGOD (GIN)} & 1.0716          & \textbf{1.0261}   & 1.0215                              & \textbf{1.0655} \\
{VGOD (GCN)} & \textbf{1.0637} & 1.0278            & 1.0214                              & 1.0713          \\
{VGOD (GAT)} & 1.0680          & 1.0268            &{\textbf{1.0211}}& 1.0672     \\ \hline   
\end{tabular}
\vspace{-2mm}
\end{table}


\subsubsection{Labeled outlier study}
\begin{figure}[!htbp]
	 \vspace{-2mm}
		\setlength{\belowcaptionskip}{-0.5cm}   
		\centering
		\includegraphics[width=200.0 pt]{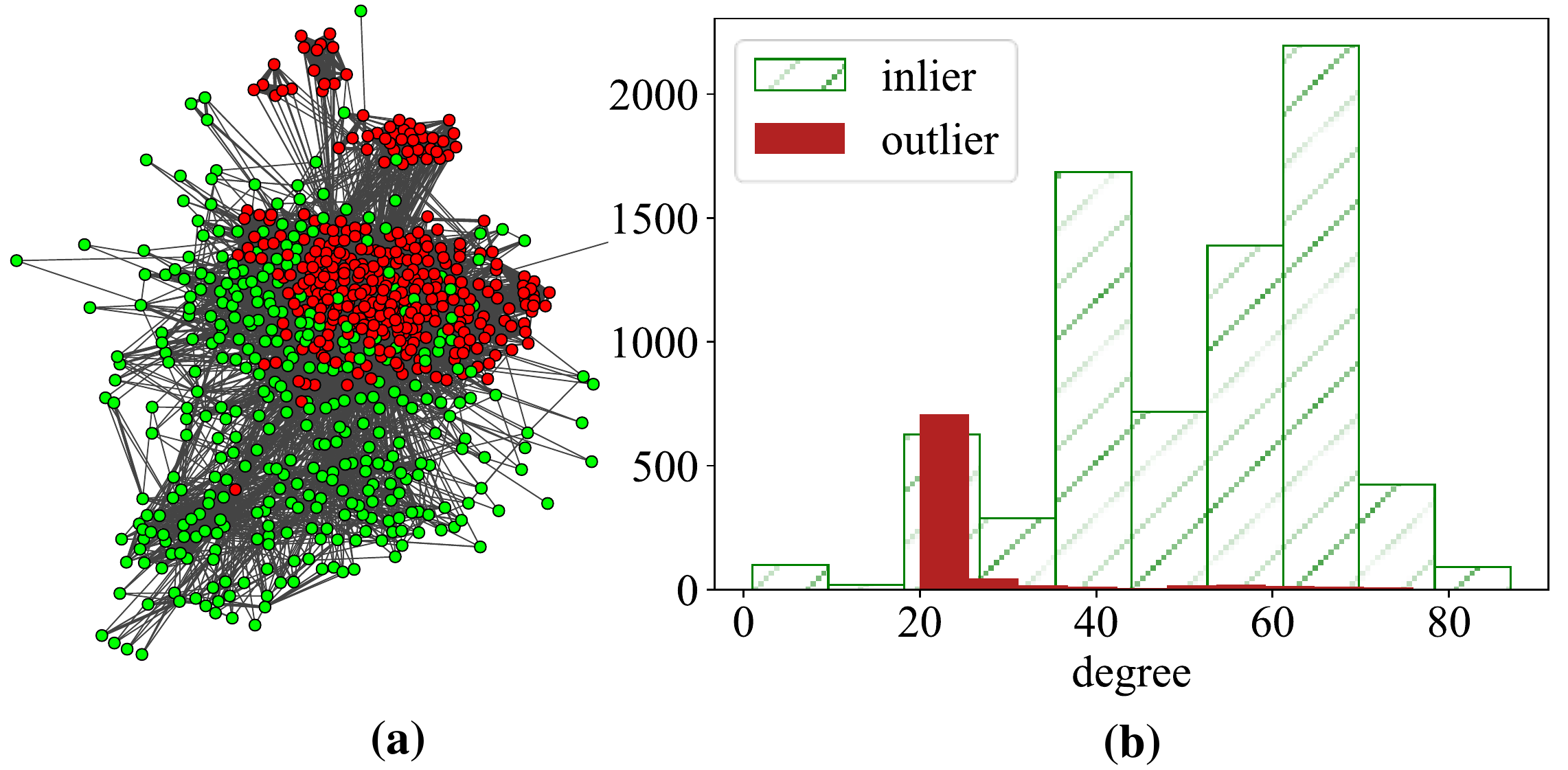}
  \vspace{-3mm}
		\caption{(a) A subgraph in Weibo. The nodes in red denote the outliers. (b) Node degree distribution of Weibo.}%
		\label{weibo-g}
		 \vspace{-2mm}
\end{figure}

\newtext{
We make the analysis  on the Weibo dataset. We compare VGOD with the second best baseline AnomalyDAE in Table \ref{Weibo-analysis}. It reveals that the main reason for VGOD's superior performance is its improvement in structural outlier detection. It is shown in Fig \ref{weibo-g}(b) that outliers do not have a higher node degree distribution.  In addition, we find that attribute vectors of outliers are more diverse, as the variance of attribute vectors among all outliers is 425.0 and that of the inliers is 11.95.}

\newtext{
From Fig \ref{weibo-g}(a), we find that both inliers (green points) and outliers (red points) are quite cohesive. The homophily \cite{homophily-from} of the whole graph is 0.75. Note that a random graph has a homophily of 0. 
In this case, these outliers, which differ greatly from each other, are connected closely, forming clusters of structural outliers. Therefore, it is easily detected by the neighbor variance of VGOD. There are also a lot of clusters formed by inliers. They are not regarded as outliers since their attributes are closed.
}

\begin{table}[!htbp]
\vspace{-3mm}
  \centering
  \caption{AUC detail in the Weibo}
  \renewcommand{\arraystretch}{1.1} 
    \label{Weibo-analysis}
    \begin{tabular}{c|ccc}
    \toprule
          & \textbf{$AUC$} & \textbf{$AUC(\mathcal{V}^-,\mathcal{O}^{str})$} & \textbf{$AUC(\mathcal{V}^-,\mathcal{O}^{attr})$} \\
    \midrule
    \textbf{VGOD} & 0.977  & 0.922 & 0.926 \\
    \textbf{AnomalyDAE} & 0.925 & 0.796 & 0.925 \\
    \bottomrule
    \end{tabular}%
  \label{tab:addlabel}%
  \vspace{-2mm}
\end{table}%

\subsubsection{Effect of the self-loop edge}
\label{eff-self-loop}
\newtext{We study the effect of the self-loop edge technique, which enables  neighbor variance to detect contextual outliers besides structural outliers.
In the first step, we study the effect of variance-based model to detect contextual outliers. We inject only contextual outliers into the datasets, using the same injection parameters as Section \ref{leak-inject-setting}.  In Table \ref{tab: nv-for-attr} and Table \ref{tab:vgod-sl}, ``w/ SL''  means employing the self-loop edge technique. The result in Table \ref{tab: nv-for-attr} shows that neighbor variance with this simple technique indeed has an effect on detecting contextual outliers, especially in those citation networks with small node degrees. }

\vspace{-3mm}
\begin{table}[!htbp]
  \centering
  \caption{AUC of VBM to detect contextual outliers}
  \label{tab: nv-for-attr}
    \begin{tabular}{l|llll}
    \toprule
          & \multicolumn{1}{l}{Cora} & \multicolumn{1}{l}{Citeseer} & \multicolumn{1}{l}{PubMed} & \multicolumn{1}{l}{Flickr} \\
    \midrule
    VBM   & 0.5026 & 0.5128 & 0.4883 & 0.4725 \\
    VBM w/ SL & 0.7978 $\uparrow$ & 0.8567 $\uparrow$ & 0.8364 $\uparrow$ & 0.6463 $\uparrow$ \\
    \bottomrule
    \end{tabular}%
  \label{tab:addlabel}%
\end{table}%

\begin{table}[!htbp]
\vspace{-4mm}
  \centering
  \caption{AUC of VGOD in ablation of self-loop edge}
  \label{tab:vgod-sl}
    \begin{tabular}{l|lllll}
    \toprule
          & Cora  & Citeseer & PubMed & Flickr & Weibo \\
    \midrule
    VGOD  & 0.8911 & 0.9485 & 0.9592 & 0.8773  & 0.9707 \\
    VGOD w/ SL & 0.9503 $\uparrow$  & 0.9845 $\uparrow$  & 0.9813 $\uparrow$  & 0.8313 & 0.9765 $\uparrow$  \\
    \bottomrule
    \end{tabular}%
  \label{tab:addlabel}%
\end{table}%
\vspace{-0.5mm}

\newtext{In the second step, we do the ablation study of this technique under the UNOD experiment.
Table \ref{tab:vgod-sl} demonstrates that self-loop edge also greatly improves the detection performance of VGOD in those citation networks due to the extra utilities of neighbor variance on contextual outlier detection.}

\section{Conclusion}
In this paper, we revisit the problem of unsupervised node outlier detection. Firstly, we find that the current outlier injection approach exists a serious data leakage issue and make a theoretical analysis in depth. Secondly, we propose a new framework, which consists of a novel variance-based model and a more general attribute reconstruction model to detect two types of outliers. Our model successfully outperforms all previous SOTA models with the best outlier detection performance and  the detection balance. 

We believe our insight into the data leakage issue will lead to better outlier injection approaches and algorithms for UNOD. Moreover, the concept of neighbor variance may also exhibit  great potential in other research areas such as graph mining and graph representation learning in the future. 

\vspace{12pt}

\small
\bibliographystyle{IEEEtran}
\bibliography{ref}

\begin{appendices}
\hypertarget{Appendix}{}
\section{Additional experiments}

\subsection{Effect of score combination Strategy}
We study the effectiveness of mean-std normalization employed in VGOD. We replace the score combination strategy with weighted sum and  ``sum-to-unit'' normalization on the UNOD experiment.
Normalization of ``sum-to-unit'' can be formalized as:
\begin{equation}
\small
     \begin{split}
        \widehat{o_i}^{(k)} & = \frac{o_i^{(k)}}{\sum_{j \in |\mathcal{V}|} o_j^{(k)}} \\
    o_i & =  \sum_{k \in K} \widehat{o_i}^{(k)}
    \end{split}
    \label{eq:sum-to-unit}
    \normalsize
\end{equation} 
where $o_i^{(k)}$ means the $k^{th}$ outlier score of node $v_i$ outputted by model and  $o_i^{(k)}$ in Eq. (\ref{eq:sum-to-unit}) should be greater than 0.

\begin{table}[htbp]
  \centering
  \caption{AUC for different score combination on the UNOD experiement}
  \label{tab:score-auc}
    \begin{tabular}{c|ccccc}
    \toprule
    Model & Cora  & Citeseer & PubMed & Flickr & Weibo \\
    \midrule
    VGOD (mean-std) & \textbf{0.956 } & \textbf{0.987 } & 0.981  & \textbf{0.883 } & \textbf{0.976 } \\
    VGOD (weight) & 0.919  & 0.859  & \textbf{0.982 } & 0.729  & 0.942  \\
    VGOD (sum-to-unit) & 0.935  & 0.957  & 0.981  & 0.850  & 0.970  \\
    \bottomrule
    \end{tabular}%
  \label{tab:addlabel}%
\end{table}%

\begin{table}[htbp]
  \centering
  \caption{AUCGAP for different score combination on the UNOD experiement}
  \label{tab:score-gap}
    \begin{tabular}{c|cccc}
    \toprule
    Model & Cora  & Citeseer & PubMed & Flickr \\
    \midrule
    VGOD (mean-std) & \textbf{1.0680 } & \textbf{1.0268 } & 1.0211  & \textbf{1.0672 } \\
    VGOD (weight) & 1.0781  & 1.3641  & 1.0095  & 1.9662  \\
    VGOD (sum-to-unit) & 1.1716  & 1.1133  & \textbf{1.0000 } & 1.2241  \\
    \bottomrule
    \end{tabular}%
  \label{tab:addlabel}%
\end{table}%

In Table \ref{tab:score-auc} and Table \ref{tab:score-gap}, 
both the AUC and \textit{AucGap} scores of VGOD with mean-std normalization are significantly superior to others on Cora, Citeseer, Flickr and only slightly inferior to the highest score on PubMed.


\subsection{Effect of Inductive Inference}
We extend the UNOD experiment to the inductive setting here. First, we use the datasets in the UNOD experiment to train all algorithms in Section \ref{UNOD-exp}. Then, we inject outliers with a different random seed but the same approach to generate a new group of datasets for evaluation of all algorithms. Other settings are consistent with the UNOD experiment. Note that AnomalyDAE cannot perform inductive inference.

\begin{table}[htbp]
  \centering
  \caption{Auc for the UNOD experiment in the inductive setting}
  \label{tab:induct-auc}
    \begin{tabular}{c|cccc}
    \toprule
    Model & Cora  & Citeseer & PubMed & Flickr \\
    \midrule
    Dominant & 0.8531  & 0.8755  & 0.8089  & 0.7545  \\
    DONE  & \underline{0.9110}  & \underline{0.9545}  & 0.8362  & \underline{0.7794}  \\
    CoLA  & 0.7698  & 0.8133  & 0.9076  & 0.6570  \\
    CONAD & 0.7139  & 0.7074  & 0.6817  & 0.7536  \\
    \midrule
    DegNorm & 0.8873  & 0.9350  & \underline{0.9120}  & 0.7642  \\
    VGOD  & \textbf{0.9693 } & \textbf{0.9840 } & \textbf{0.9783 } & \textbf{0.8977 } \\
    \bottomrule
    \end{tabular}%
  \label{tab:addlabel}%
  \vspace{-2mm}
\end{table}%


\begin{table*}[htbp]
  \centering
  \vspace{-2mm}
    \caption{AucGap for the UNOD experiment in the inductive setting}
   \label{tab:induct-aucgap}
    \begin{tabular}{c|ccc|ccc|ccc|ccc}
    \toprule
    \multirow{2}[2]{*}{Model} & \multicolumn{3}{c|}{Cora} & \multicolumn{3}{c|}{Citeseer} & \multicolumn{3}{c|}{PubMed} & \multicolumn{3}{c}{Flickr} \\
          & aucgap & str   & context & aucgap & str   & context & aucgap & str   & context & aucgap & str   & context \\
    \midrule
    Dominant & 1.379  & 0.709  & 0.977  & 1.286  & 0.758  & 0.975  & 1.617  & \textcolor[rgb]{ 1,  0,  0}{\textit{0.615}} & 0.994  & 1.961  & \textcolor[rgb]{ 1,  0,  0}{\textit{0.504}} & 0.989  \\
    DONE  & 1.223  & 0.990  & 0.809  & 1.116  & 0.996  & 0.892  & 1.302  & 0.940  & 0.722  & 1.701  & \textcolor[rgb]{ 1,  0,  0}{\textit{0.571}} & 0.971  \\
    CoLA  & \underline{1.058}  & 0.741  & 0.784  & 1.246  & 0.894  & 0.718  & 1.102  & 0.945  & 0.857  & \underline{1.243}  & \textcolor[rgb]{ 1,  0,  0}{\textit{0.582}} & 0.723  \\
    CONAD & 2.030  & \textcolor[rgb]{ 1,  0,  0}{\textit{0.467}} & 0.948  & 2.245  & \textcolor[rgb]{ 1,  0,  0}{\textit{0.433}} & 0.972  & 2.578  & \textcolor[rgb]{ 1,  0,  0}{\textit{0.379}} & 0.978  & 1.968  & \textcolor[rgb]{ 1,  0,  0}{\textit{0.503}} & 0.989  \\
    \midrule
    DegNorm & 1.191  & 0.953  & 0.800  & \underline{1.104}  & 0.971  & 0.879  & \underline{1.099}  & 0.863  & 0.948  & 1.759  & \textcolor[rgb]{ 1,  0,  0}{\textit{0.548}} & 0.964  \\
    VGOD  & \textbf{1.020} & 0.965  & 0.946  & \textbf{1.000} & 0.973  & 0.973  & \textbf{1.021} & 0.961  & 0.981  & \textbf{1.033} & 0.871  & 0.900  \\
    \bottomrule
    \end{tabular}%
  \label{tab:addlabel}%
\end{table*}%

Table \ref{tab:induct-auc} and \ref{tab:induct-aucgap} show the AUC and \textit{AucGap} scores for these baselines and our VGOD. The result is similar to the transductive setting, where our VGOD outperforms other baselines significantly. The performance of our VGOD is even better than  the performance in the transductive setting. The reason is owing to the case that the overfitting of  issue  is eliminated in the inductive setting.

\end{appendices}

\end{document}